AOS-0001

# The Agent Operating System (AOS): A Reference Operating Architecture for Distributed Agentic Systems

*A Vendor-Neutral Architectural Foundation for Governable, Reliable, and Interoperable Agentic AI Systems*

**Ankur Sharma**

*Lead Author, AOS Specification Series*

**Deep Shah**

*Co-Author, AOS Specification Series*

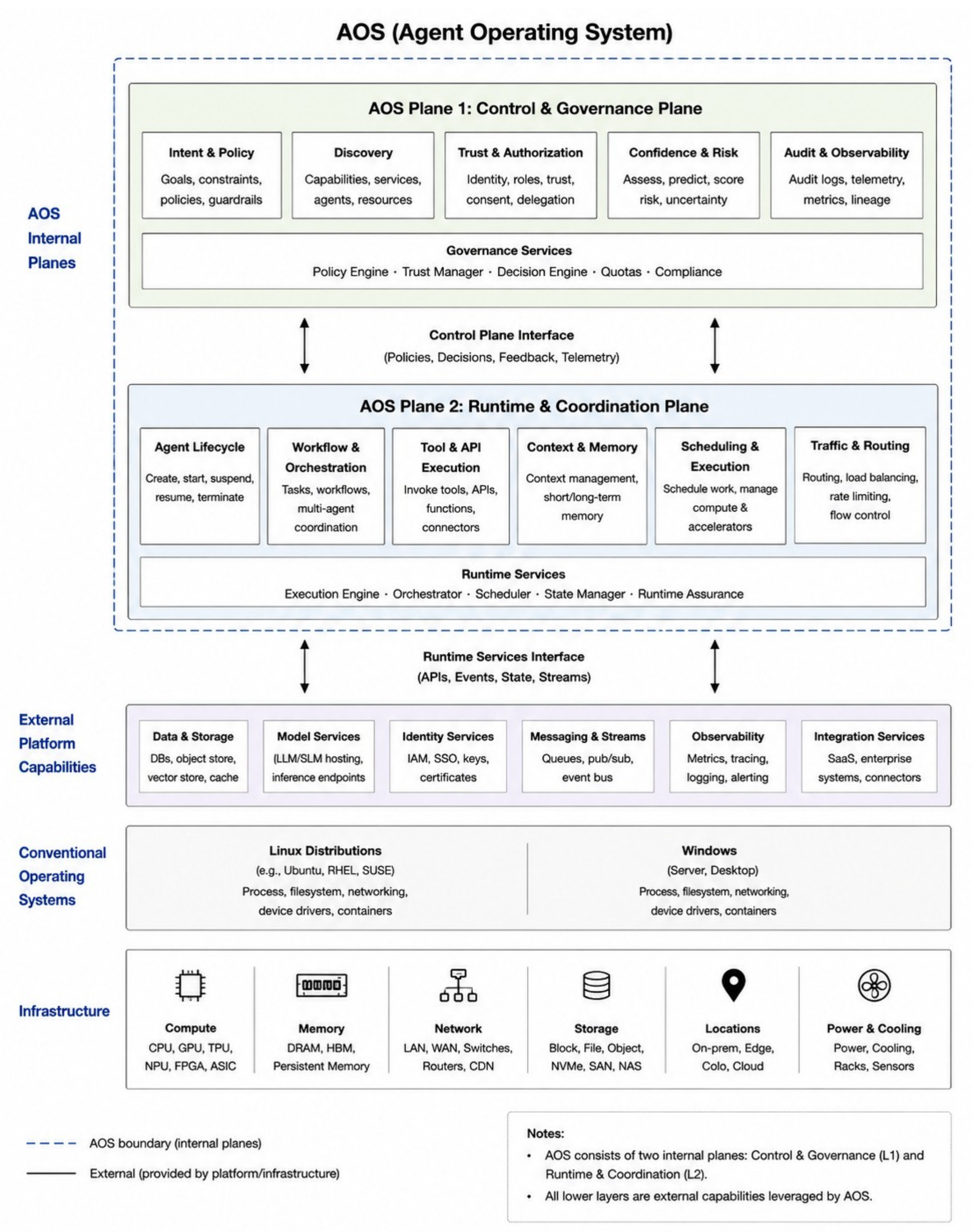

*Conceptual cover figure: the AOS boundary encloses the Control & Governance Plane and the Runtime & Coordination Plane; platform services, conventional operating systems, and infrastructure remain external dependencies.*

**Status.** This manuscript is a research and architecture proposal. It defines the foundational boundaries of AOS and a program of subsequent specifications. It does not claim benchmark superiority or universal completeness.

# Document Control and Editorial Model

*Table 1. Document control and editorial metadata.*

| Field | Value |
| --- | --- |
| **Specification identifier** | AOS-0001 |
| **Document type** | Reference architecture and systems research preprint |
| **Status** | Draft v0.8 |
| **Authors** | Ankur Sharma (Lead Author), Deep Shah (Co-Author) |
| **Relationship to AOS-0000** | The initial AOS concept was published by Sharma and Shah as arXiv:2606.01508 [1]. AOS-0001 extends that work into the present two-plane reference operating architecture. |
| **Relationship to aos-core** | aos-core is an evolving reference implementation. It validates selected architectural concepts but does not define the architecture. |
| **Contribution model** | Review comments are acknowledged. Adopted text, figures, algorithms, or implementation artifacts are credited. Co-authorship requires substantial intellectual contribution. |
| **Submission target** | arXiv preprint; subsequent IEEE, ACM, OCP, or community specification review may follow. |

## Normative and Research Language

The terms MUST, MUST NOT, REQUIRED, SHALL, SHALL NOT, SHOULD, SHOULD NOT, RECOMMENDED, MAY, and OPTIONAL are used according to BCP 14 when they appear in uppercase [12, 13]. Because this document is both a research proposal and a foundational architecture, normative statements are limited to architectural invariants and conformance declarations. Protocol-level wire behavior is delegated to later AOS specifications.

## Editorial Position

- Architecture precedes implementation: no product, protocol, framework, or repository defines AOS by itself.
- AOS is implementation-plural: multiple conforming control planes, runtimes, registries, policy engines, and assurance mechanisms may coexist.
- The lead-editor model preserves conceptual consistency while allowing open technical review and attributed contribution.
- Planned AOS specifications may refine AOS-0001, but any contradiction requires an explicit architectural rationale, compatibility analysis, and a versioned amendment.

# Contents

# 1. Abstract

Large language models have transformed artificial intelligence from isolated prediction services into components of long-running, distributed systems that reason, invoke tools, retrieve external state, delegate tasks, and act on behalf of users and organizations. The surrounding ecosystem has responded with agent frameworks, workflow engines, model-serving platforms, memory systems, communication protocols, and observability tools. These technologies improve execution, but they do not provide a stable, implementation-independent operating architecture for governing intent, selecting capabilities, preserving authority across delegation, controlling uncertainty, coordinating runtime behavior, and reconstructing why consequential actions occurred. This paper proposes the Agent Operating System (AOS), a vendor-neutral reference operating architecture for distributed agentic systems. AOS contains two internal planes: a Control & Governance Plane responsible for intent, policy, trust, authority, confidence, auditability, observability, and human oversight; and a Runtime & Coordination Plane responsible for agent lifecycle, workflow coordination, model and tool routing, context and memory coordination, scheduling, traffic management, and runtime assurance. Platform services, Linux or Windows, container runtimes, and physical infrastructure remain outside the AOS boundary and are integrated through explicit interfaces. The paper specifies AOS concepts, invariants, interface objects, optimization objectives, deployment profiles, and reliability responsibilities. It also identifies tradeoffs and unresolved research questions. AOS is not presented as a replacement for existing frameworks or infrastructure; it is proposed as the operating architecture through which heterogeneous components can be composed into governable, reliable, observable, and interoperable agentic systems.

## 2. Executive Summary

Artificial intelligence is undergoing a systems transition. A modern agentic application is rarely a single model call. It may contain a planner, specialized agents, model endpoints, retrieval systems, durable memory, tool adapters, business workflows, human approval points, identity services, policy engines, and geographically distributed compute. The resulting system is dynamic: behavior is generated at runtime, execution paths vary with context, providers are selected conditionally, and authority may pass through multiple delegation steps before reaching an API, process, network connection, or system call.

This transition changes the operating problem. Traditional software architecture assumes that application logic determines the execution path in advance. Agentic systems instead combine probabilistic reasoning with deterministic external effects. The model may decide which tool to call, which sub-agent to consult, what information to retrieve, or whether to revise the plan. The operational environment must therefore govern not only resources but also semantics: what objective is being pursued, which actor has authority, which capability is eligible, which constraints remain binding, what uncertainty exists, and what evidence justifies continuation.

Most current agent stacks are execution-centric. Agent frameworks provide reasoning loops and tool invocation. Workflow engines manage state transitions. Model servers perform inference. Retrieval systems expose knowledge. Kubernetes schedules containers. Linux and Windows control processes, files, memory, sockets, and devices. OpenTelemetry and related systems collect traces, metrics, logs, and baggage [9, 10]. MCP standardizes model-facing access to tools and context [7], while A2A standardizes communication among opaque agentic applications [8]. Each is useful; none by itself defines an end-to-end operating model for distributed agency.

AOS addresses this architectural gap through a two-plane design. The Control & Governance Plane answers what should happen, under whose authority, and subject to which constraints. It captures and normalizes intent; discovers eligible capabilities; evaluates policy, trust, consent, and delegation; derives confidence actions; and records audit and observability obligations. The Runtime & Coordination Plane answers how approved work should be coordinated and assured. It manages agent lifecycle, workflow progress, model and tool routing, context and memory coordination, scheduling, traffic management, protocol mediation, and runtime assurance.

The separation is logical rather than necessarily physical. A small AOS implementation may place both planes inside one process. A regional deployment may distribute them across services. A federated deployment may maintain local control domains with shared capability and policy federation. In every case, the architecture requires explicit state ownership and traceable interfaces between decisions and actions. This separation allows control semantics to evolve independently from runtime technologies, and it prevents a particular workflow engine, model provider, cloud platform, or operating system from becoming the de facto definition of AOS.

AOS is capability-centric. Consumers request a stable capability, such as ai.case.summary.generate, rather than binding directly to a model, agent, or vendor. Multiple providers may implement the capability. The control plane first rejects providers that violate hard constraints, such as authorization, data residency, protocol compatibility, or minimum trust. It may then rank the feasible candidates according to semantic fit, reliability, latency, cost, locality, energy, and confidence. Runtime coordination binds the selected provider and invokes it through a protocol adapter. If output verification fails, the control loop may retry, select a fallback, route to another region, escalate to a human, or terminate.

Authority and delegation are first-class. Authentication identifies an actor; authority defines what the actor may do. When an agent delegates to another agent or tool, the delegated scope must not exceed the parent scope, and the lineage must remain observable. This requirement responds to an emerging observability gap: conventional traces may record all calls yet remain unable to reconstruct which actions occurred under a particular delegation [4]. AOS therefore propagates delegation identifiers, parent relationships, root intent, policy obligations, confidence thresholds, and trace context with execution.

Confidence is also elevated from a dashboard value to a control signal. AOS does not assume that a model probability is equivalent to operational confidence. Instead, confidence may incorporate intent clarity, data quality, model calibration, provider trust, policy certainty, temporal freshness, validation results, and runtime evidence. A confidence decision maps the current evidence to an action such as proceed, proceed with warning, retry, fallback, escalate, or terminate. The planned AOS-0005 document is intended to define the detailed aggregation and propagation rules; this paper establishes the architectural requirement that confidence remain evidence-backed and actionable.

The architecture is deliberately compatible with existing systems. Platform services provide models, data, identity, messaging, policy decision points, and telemetry backends. Linux and Windows remain responsible for conventional process and resource control. Lower-level assurance may translate an approved authority chain into process, filesystem, network, namespace, cgroup, or syscall constraints. The Linux kernel, for example, exposes mechanisms such as cgroup v2 for hierarchical resource control [25] and seccomp for syscall filtering [24]. These mechanisms are enforcement substrates; they do not replace the semantic control plane.

The intended outcome is not a monolithic product. AOS-0001 supplies the foundational architecture; planned follow-on specifications AOS-0002 through AOS-0010, summarized in Appendix G, are intended to refine plane boundaries, capability discovery, confidence, reliability, protocol mappings, the ecosystem landscape, the maturity model, and the aos-core reference implementation. With the exception of the previously published AOS work cited as [1], these documents are planned and have not yet been published.

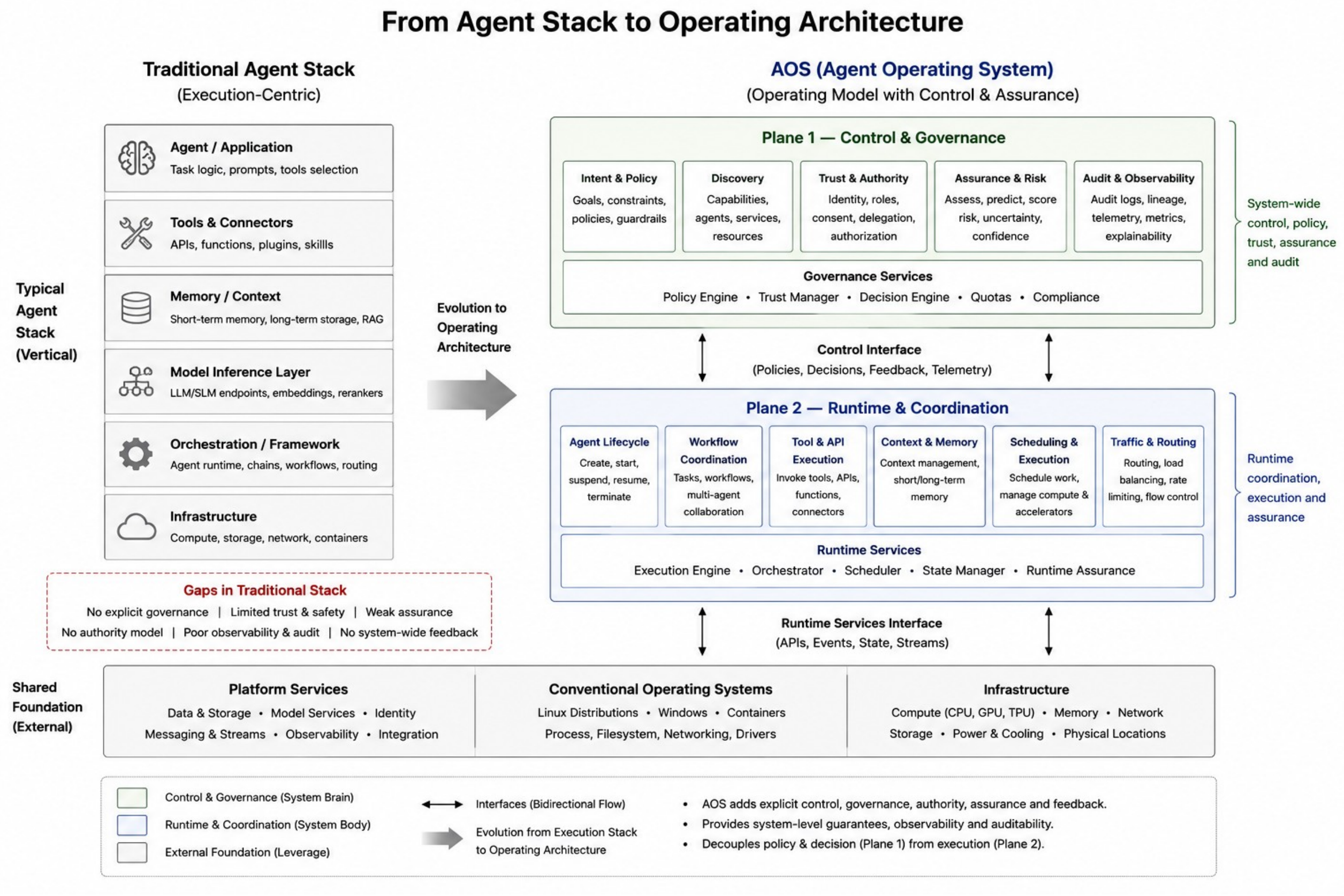

*Figure 1. From an execution-centric agent stack to an operating architecture with explicit governance, authority, assurance, and feedback.*

# 3. Problem Statement

## 3.1 From Model Invocation to Distributed Agency

The computing industry repeatedly introduces new abstraction layers when the managed system becomes too complex for the previous layer. Operating systems abstracted devices and protected concurrent programs. Virtualization decoupled operating environments from physical machines. Cloud platforms exposed elastic resources. Containers standardized packaging, and Kubernetes introduced declarative orchestration and reconciliation for fleets of workloads [11]. Agentic AI introduces another transition: the unit being coordinated is no longer only a process, container, request, or workflow step, but a goal-directed entity whose execution path is partially generated at runtime.

This distinction is operationally consequential. A process executes instructions supplied by a program. An agent may construct a plan, choose a model, request context, invoke a tool, delegate a subtask, revise its assumptions, and produce an external action. The runtime can therefore change the system graph while the system is active. The number of possible paths grows with the available capabilities, and the correct path depends on semantic, policy, temporal, and environmental state. Conventional infrastructure remains necessary, but it is insufficient to represent these relationships.

Agentic systems also cross organizational and geographic boundaries. An intent created by a human in one trust domain may be interpreted by a control service, delegated to an agent hosted by a second organization, routed to a model in a third region, and completed through an enterprise API. Every transition can alter risk, latency, cost, data exposure, and accountability. The architecture must preserve the connection between the originating objective and the final side effect even when the implementation spans heterogeneous protocols and providers.

## 3.2 The Execution-Centric Fragmentation Problem

The current ecosystem is rich in specialized components but fragmented in responsibility. Agent frameworks define local reasoning and orchestration. Model platforms expose inference and evaluation. Tool protocols improve integration. Observability systems collect telemetry. Governance frameworks articulate desired properties. Security products enforce selected controls. The fragmentation is not a defect in any individual component; it is a consequence of independent evolution. The systems problem appears when an enterprise attempts to combine them into one operational environment.

Without an explicit operating architecture, every application invents its own mapping from intent to policy, capability, provider, and execution. Trust may be represented as an identity claim in one component, a vendor allowlist in another, and a score in a third. Confidence may be a model logit, an evaluation score, or an operator judgment. Delegation may be represented only by nested requests. Audit logs may show calls without recording why a provider was selected or which authority permitted the call. As systems scale, these local conventions become incompatible and difficult to govern.

Execution-centric architectures also encourage late governance. The reasoning graph is constructed first and a guardrail is inserted near the output. This approach can catch some unsafe text but cannot reliably govern data access, tool choice, provider location, delegation scope, token budget, process creation, or network behavior. By the time an output filter runs, side effects may already have occurred. The architecture must move policy, trust, authority, and assurance into the lifecycle rather than applying them only after generation.

## 3.3 The Semantic Control Gap

Production agentic systems require a semantic control layer that can answer questions not represented by conventional runtime interfaces:

- What objective is authoritative, and how is ambiguity represented?
- Which actor or organization has authority to request, approve, delegate, or revoke an action?
- Which capability is required, and which providers are feasible under policy, trust, locality, cost, and performance constraints?
- What confidence exists before and after execution, and what control action follows from insufficient confidence?
- Which context and memory records influenced the decision, and what is their provenance and freshness?
- What state transitions are legal, and how should the system recover from partial failure, contradiction, or provider loss?
- How can an operator reconstruct the causal and delegation chain from human objective to external effect?

These questions are semantic because their answers depend on meaning, authority, and policy, not only resource availability. They are operational because the answers must influence execution in real time. AOS therefore treats intent, capability, authority, delegation, confidence, and evidence as managed system objects rather than incidental application metadata.

The control gap is especially visible when agents can act. An apparently benign request such as “investigate the failed deployment” may produce file reads, process creation, network connections, credential use, and configuration changes. At the planning layer, the action is described in human-scale terms. At the operating-system layer, the action becomes syscalls and resource access. A robust architecture must preserve an authority chain across this semantic distance and provide deterministic enforcement boundaries before consequential effects.

## 3.4 Central Research Question

**Research question.** What implementation-independent operating architecture is required to make distributed agentic systems governable, reliable, observable, trustworthy, and interoperable while allowing heterogeneous runtimes, protocols, platforms, operating systems, and infrastructure to evolve independently?

The question is architectural rather than product-specific. A valid answer must remain useful if a model provider changes, a workflow engine is replaced, a new protocol appears, or workloads move from cloud to edge. It must define stable responsibilities and interfaces without assuming that one organization owns the complete system.

The paper answers by proposing a two-plane AOS boundary and by formalizing the objects and invariants that couple control to execution. The proposal is evaluated conceptually against existing systems, distributed-systems constraints, deployment profiles, and failure modes. The current work does not claim a universal optimal scheduler or a complete protocol; those are intentionally delegated to later specifications and empirical work.

## 3.5 Scope, Assumptions, and Non-Goals

*Table 2. Scope and non-goals.*

| Category | Position in AOS-0001 |
|---|---|
| **In scope** | Architectural responsibilities, plane boundaries, managed objects, interfaces, invariants, deployment profiles, mathematical objectives, and conformance declarations. |
| **Assumption** | Agentic systems combine probabilistic decision components with deterministic external effects and may span multiple runtimes and administrative domains. |
| **Assumption** | Existing operating systems and platform services remain present and provide lower-level execution and enforcement. |
| **Non-goal** | Defining a universal agent programming language, prompt format, workflow DSL, model API, memory database, or inference engine. |
| **Non-goal** | Mandating a specific cloud provider, model vendor, policy engine, identity system, observability backend, or kernel mechanism. |
| **Non-goal** | Claiming benchmark superiority. Quantitative examples in this paper illustrate architectural reasoning and are not empirical product results. |

# 4. Contributions

*Table 3. Summary of contributions.*

| Contribution | Summary |
|---|---|
| **A vendor-neutral reference operating architecture** | The paper defines a stable AOS boundary independent of runtime framework, protocol, operating system, cloud, or model provider. |
| **Two-plane separation of governance and execution** | The Control & Governance Plane and the Runtime & Coordination Plane have distinct state ownership and communicate through explicit directives, evidence, and feedback. |
| **Capability-centric abstraction** | A stable capability name separates the requested outcome from replaceable providers such as agents, models, tools, workflows, humans, or services. |
| **Authority-preserving delegation** | Delegation is modeled as bounded transfer of authority with parent lineage, scope, expiry, obligations, and revocation semantics. |
| **Confidence as an operational control signal** | Confidence becomes evidence-backed state that can trigger proceed, warning, retry, fallback, escalation, or termination. |
| **Semantic observability and provenance** | The architecture augments logs, metrics, and traces with intent, capability, policy, trust, confidence, context, delegation, and causal relationships. |
| **Distributed and multi-region operating model** | AOS supports embedded, centralized, regional, federated, edge, and sovereign deployments while preserving local authority and global interoperability. |
| **Mathematical formulation** | The paper formulates feasibility constraints, provider ranking, confidence aggregation, authority monotonicity, placement, latency, cost, energy, risk, and reliability objectives. |
| **Conformance and extension model** | Canonical objects, service-provider interfaces, protocol bindings, and profiles allow multiple implementations to claim bounded and testable alignment. |
| **Foundational basis for the AOS series** | The paper establishes the core architectural invariants that later specifications may refine but not silently contradict. |

The contributions are intentionally architectural. They create a vocabulary and a set of invariants against which implementations can be compared. They do not require the community to converge on one source code base. The aos-core project provides one reference implementation path, but alternative implementations may satisfy the same responsibilities through different designs.

The paper also contributes negative scope. It explicitly identifies what AOS does not replace. This is important because the term “AI operating system” is used for model platforms, agent runtimes, orchestration products, data platforms, and infrastructure schedulers. A reference architecture is valuable only if it can classify these systems without absorbing every capability into an unbounded definition.

# 5. Related Work

AOS draws from operating systems, distributed systems, cloud orchestration, agent frameworks, AI governance, observability, security, and assurance. The proposal is complementary to these areas. This section identifies the responsibilities they solve and the architectural gap that remains when they are composed into distributed agentic systems.

## 5.1 Traditional Operating Systems and Security Boundaries

Operating systems provide the foundational abstractions for process lifecycle, scheduling, virtual memory, filesystems, networking, identity, access control, and device management [20, 21]. Their key contribution is mediation: applications do not manipulate hardware directly, and protection mechanisms prevent one workload from arbitrarily consuming or modifying another workload's resources. This principle remains central to AOS. Consequential agent actions must eventually be mediated by a lower layer that the workload cannot bypass.

The analogy, however, has limits. A Linux or Windows kernel does not natively understand that a process exists to satisfy a customer-support intent, that a network connection was created under delegated authority, or that a model output is below an operational confidence threshold. It sees processes, credentials, files, sockets, memory pages, and syscalls. AOS therefore adds semantic control above the conventional operating system while preserving the latter as the final local enforcement substrate.

Modern kernels expose mechanisms that can implement AOS obligations. Linux cgroup v2 organizes processes hierarchically and distributes resources through controllers [25]. Seccomp BPF can restrict a process to a reduced syscall surface [24]. Namespaces, capabilities, LSMs, container runtimes, and hypervisors provide additional isolation. These mechanisms implement resource and action constraints; the AOS control plane supplies the high-level authority and policy that justify those constraints.

## 5.2 Distributed Systems, Cluster Managers, and Reconciliation

Distributed systems research established the importance of explicit failure models, partial synchrony, causal order, consensus, replication, and consistency [14, 15, 44, 45]. AOS inherits the assumption that messages can be delayed, components can fail independently, clocks are imperfect, and state can become stale. Accordingly, the architecture does not assume a single omniscient controller. Regional and federated domains must tolerate partition and conflicting local observations and support eventual reconciliation.

Cluster managers such as Borg, Mesos, Omega, and Kubernetes separate desired state from worker execution and provide scheduling, service discovery, controllers, and extensibility [11, 39–41]. Kubernetes in particular distinguishes a control plane that manages global cluster state from node components that execute workloads [11]. Distributed execution engines such as Ray and Naiad similarly separate coordination from distributed task execution for demanding AI and dataflow workloads [37, 38]. AOS adopts the control-plane pattern but changes the managed objects and decision criteria. Pods are scheduled primarily according to resource and placement constraints; AOS must additionally reason about semantic capability, authority, trust, confidence, policy, data residency, human approval, and evidence obligations.

AOS is therefore not "Kubernetes for prompts." The relevant inheritance is architectural: declarative state, reconciliation, explicit APIs, pluggable providers, and separation of decisions from execution. The difference is the semantic resource model and the need to govern probabilistic, delegating actors.

## 5.3 Agent Frameworks and Workflow Engines

Agent frameworks such as AutoGen, LangGraph [35], CrewAI [36], and vendor SDKs provide mechanisms for defining agents, tools, roles, conversations, and workflows. AutoGen, for example, introduced a multi-agent conversation framework in which agents can combine LLMs, humans, and tools [34]. ReAct interleaves reasoning and actions, and Toolformer explores model-mediated tool use [32, 33]. These systems are important runtime substrates because they make dynamic execution practical.

Their primary unit is usually an application graph or agent team. Governance, trust, capability registration, confidence, and audit semantics are often implemented as application-specific extensions. A workflow can encode allowed transitions, but it does not by itself define a federated authority model or a vendor-neutral capability contract. AOS places these frameworks inside the Runtime & Coordination Plane or treats them as providers behind its interfaces.

Structured workflows remain essential. The claim is not that AOS replaces finite-state machines or orchestration. Rather, AOS makes the decision to select, constrain, and observe a workflow part of a larger operating context. The same capability may be realized by LangGraph, a conventional workflow engine, a model endpoint, or a human service without changing the consumer-facing intent.

Earlier generations of agent and service infrastructure anticipated parts of this design. FIPA agent platforms standardized agent lifecycle management and directory-based capability discovery [65], and service-oriented registries such as UDDI cataloged providers for dynamic binding [66]. AOS revisits these ideas under changed assumptions: the coordinated actors are probabilistic and construct plans at runtime, authority must be delegated and attenuated across administrative boundaries, and selection is governed by policy, trust, confidence, and evidence rather than interface matching alone.

### 5.4 AIOS and Agent Operating-System Research

AIOS research has explored kernel-like services for LLM agents. Mei et al. isolate LLM, tool, scheduling, context, memory, storage, and access-control services in an AIOS kernel and report improved execution efficiency for concurrent agents [2]. This work is closely aligned with the Runtime & Coordination Plane, especially agent scheduling, context management, resource isolation, and tool access.

The earlier AOS paper by Sharma and Shah proposed an agentic control plane integrated with or positioned above conventional operating systems [1]. The present paper extends that concept into a two-plane architecture, formalizes the AOS boundary, and separates external platform services from AOS responsibilities. TopoClaw adds human-centric, topology-aware cross-device and cross-user execution, demonstrating the importance of physical and social topology, provenance, and distributed authority [3].

These efforts illustrate why the term “agent operating system” needs a reference architecture. Some systems emphasize single-host runtime kernels; others emphasize orchestration, multi-device action placement, data platforms, or governance. AOS-0001 does not declare one interpretation invalid. It provides a layered model in which each implementation can state which responsibilities it supplies.

### 5.5 Model Serving, Data Platforms, and AI Clouds

Model-serving systems and AI platforms provide deployment, scaling, routing, retrieval, experiment tracking, data processing, and lifecycle services. They are often marketed as comprehensive AI operating environments. Their breadth makes them valuable platforms, but the architecture is typically coupled to a vendor, data plane, or cloud control model. AOS treats these platforms as external capability providers and platform services.

This separation is analogous to the difference between a reference architecture and a distribution. A Linux distribution implements an operating-system architecture but does not redefine every operating-system concept. Similarly, an AI data platform may implement model routing, governance, observability, and workflows while remaining one possible realization of selected AOS responsibilities. The reference architecture enables comparison without requiring a platform to be adopted wholesale.

### 5.6 Agent Communication and Tool Protocols

MCP standardizes how applications expose resources, prompts, and tools to model-centric clients through a JSON-RPC-based protocol with lifecycle and capability negotiation [7]. The current published specification also includes authorization, elicitation, cancellation, progress, error reporting, and logging. A2A standardizes communication among opaque agents and explicitly positions itself as complementary to agent development frameworks and MCP [8].

Protocols solve message interoperability, not the complete operating problem. They do not decide which capability is permitted, whether a provider meets residency requirements, how confidence changes after a call, or which human authority must approve a side effect. AOS binds protocols to architectural interfaces. MCP may implement a tool-access adapter; A2A may implement an agent-communication adapter. The control plane remains responsible for selection, authority, policy, and evidence obligations.

### 5.7 Governance, Risk, and Assurance Frameworks

The NIST AI Risk Management Framework organizes risk-management activity through Govern, Map, Measure, and Manage functions and is intentionally voluntary and use-case agnostic [5]. The NIST Generative AI Profile extends the framework for generative systems [6]. ISO standards [54, 55] and sector-specific governance frameworks similarly define objectives such as accountability, transparency, robustness, safety, and human oversight.

These frameworks are policy and management references rather than runtime operating architectures. They specify what trustworthy practice should achieve, but they do not define how an intent, provider selection, delegation, confidence gate, runtime action, and audit record interact during execution. AOS aims to provide an executable architectural substrate through which governance requirements can be mapped to system controls and evidence.

Constitutional AI and debate-based safety research explore process structures that shape model behavior [29, 30]. Such techniques can become capability providers, policy evaluators, or validation mechanisms within AOS. The operating architecture remains responsible for deciding when they are required and how their outputs affect control actions.

### 5.8 Observability, Trace Context, and Delegation Attribution

OpenTelemetry standardizes collection, processing, and export of traces, metrics, logs, baggage, and emerging profiling signals [9]. W3C Trace Context defines portable trace identifiers and propagation fields across services [10]. Dapper and related tracing systems established the practical value of end-to-end request traces [16]. These technologies form the observability substrate for AOS.

Agentic systems require additional semantics. A span that records an API call does not necessarily identify the human objective, provider alternatives, policy decision, delegated authority, confidence state, context sources, or causal justification. Recent work shows that standard audit logs and execution traces may be structurally insufficient to reconstruct delegation-scoped execution because multiple delegation assignments can produce identical conventional observables [4]. AOS therefore defines semantic correlation objects and delegation lineage that can be carried using existing trace and baggage mechanisms.

The goal is extension rather than replacement. AOS does not define a proprietary telemetry backend. It defines the semantic data that an implementation must be able to emit and correlate.

## 5.9 Security, Identity, Provenance, and Kernel Assurance

Identity and provenance systems provide important building blocks. OAuth and OpenID Connect support delegated authorization and identity assertions [51, 52]. WebAuthn provides phishing-resistant public-key credentials for human authentication [53]. SPIFFE defines workload identities for distributed systems [23]. W3C PROV, in-toto, and SLSA provide provenance models and supply-chain evidence [26–28]. AOS treats these as trust and evidence providers rather than inventing replacement mechanisms.

The term capability also has a long lineage in capability-based security, from capability lists [61] through object-capability systems [62], the confused-deputy analysis [63], and attenuable bearer tokens such as macaroons [64]. AOS uses the word in a different sense: an AOS capability is a discoverable outcome contract, closer to a service interface than to an unforgeable authority token. The role played by security capabilities in that tradition corresponds in AOS to the delegation context, which carries bounded, attenuable authority with lineage and revocation. Multi-hop delegation among agents and tools inherits the confused-deputy risk identified in that literature; propagating explicit authority chains rather than ambient credentials is a direct response to it.

At the lower boundary, kernel and sandbox controls can enforce approved resource and syscall policies. This is consistent with the historical pattern that final enforcement moves to a layer the workload cannot bypass. AOS adds the missing translation from high-level authority and intent to enforceable runtime constraints, while acknowledging that the exact translation remains an open research problem.

## 5.10 Architectural Positioning

*Table 4. Architectural positioning of AOS relative to representative systems.*

| System or category | Primary contribution | Gap addressed by AOS |
|---|---|---|
| **Linux / Windows** | Processes, memory, files, sockets, devices | Intent, semantic capability, confidence, delegation |
| **Kubernetes** | Container scheduling, reconciliation, service abstractions | Agent semantics, authority, confidence, human oversight |
| **Agent frameworks** | Reasoning loops, workflow and tool execution | System-wide governance and vendor-neutral capability control |
| **MCP** | Model-to-tool and context interoperability | Global selection, policy, trust, reliability, and lifecycle governance |
| **A2A** | Agent-to-agent interoperability | End-to-end operating architecture and authority control |
| **OpenTelemetry / Trace Context** | Telemetry and trace propagation | Semantic intent, delegation, confidence, and decision lineage |
| **NIST AI RMF** | Risk-management objectives and organizational practice | Executable plane boundaries and runtime interfaces |
| **AIOS kernels** | Agent scheduling, context, memory, tool and resource management | Broader control, governance, federation, and ecosystem interfaces |
| **AI platforms** | Integrated data, model, serving, workflow, and governance services | Implementation-independent architecture and portability |

The comparison should not be read as a criticism of the listed systems. Each deliberately solves a narrower problem. AOS is valuable only if it can compose these systems and preserve their specialization. The central novelty is the explicit operating relationship among intent, authority, capability, policy, confidence, runtime coordination, and evidence.

The remainder of the paper formalizes that relationship and identifies the boundaries that must remain stable even as protocols and products change.

# 6. Foundational Concepts

## 6.1 Introduction

Every durable computing architecture is organized around a small set of abstractions whose meaning remains stable while implementations evolve. Conventional operating systems manage processes, threads, address spaces, files, sockets, users, devices, and scheduling classes. Distributed systems add services, messages, replicas, failure detectors, consensus, and consistency. Cloud platforms add virtual resources, declarative state, placement, elasticity, and reconciliation. Agentic systems require a corresponding semantic object model.

The need is not merely terminological. If two components use the word "agent" to mean different things, interoperability may still be possible through adapters. If they disagree about the meaning of authority, delegation, policy, confidence, or evidence, safe composition becomes substantially harder. AOS therefore defines foundational concepts as architectural objects with explicit relationships and invariants. The objects can be encoded in different schemas or protocols, but their semantic roles must remain recognizable.

The concepts in this section are intentionally independent of a particular model architecture or programming framework. They apply whether a capability is realized by a large language model, a deterministic program, a human workflow, a database query, a robotic actuator, or a composite multi-agent system.

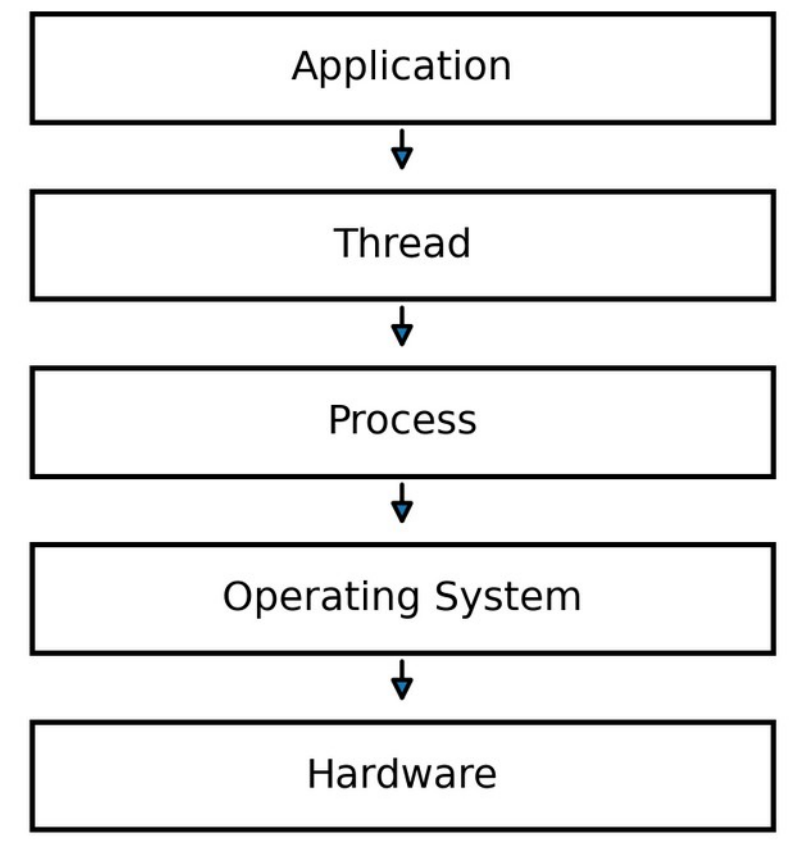

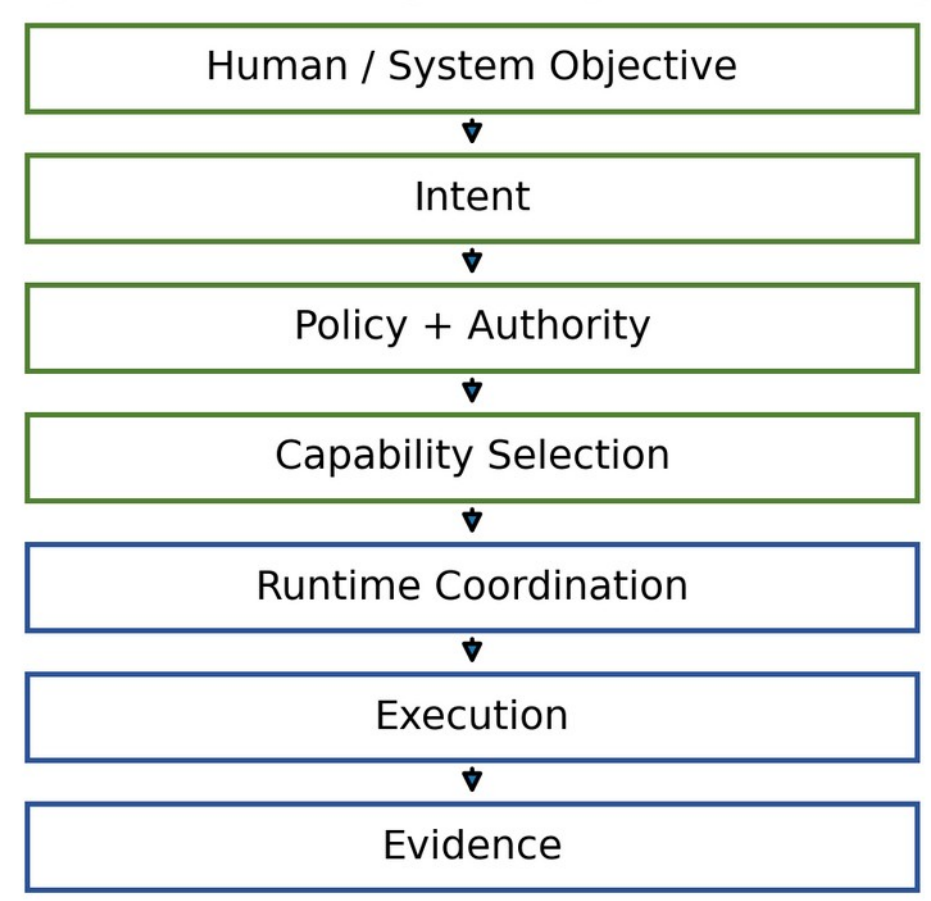

*Conventional OS manages computational resources; AOS governs semantic objectives and their realization.*

*Figure 2. Traditional computing organizes execution resources; agentic computing adds a semantic hierarchy from objective and intent to capability, execution, and evidence.*

## 6.2 Intent

### Intent

An Intent represents the highest-level semantic description of a desired outcome together with the constraints under which that outcome must be achieved. Unlike a prompt or a low-level API request, an intent does not prescribe an implementation. It describes the objective and the operating conditions while allowing AOS to choose an execution strategy.

$$I = \{G, C, A, \Pi, X, S, T\}$$

Here, G denotes the goal; C, the constraints; A, the authority; Π, the applicable policies; X, the context; S, the success criteria; and T, the temporal requirements.

Key properties:

- The original intent is immutable; revised interpretations or plans are versioned rather than silently replacing it.
- An intent is provider-independent and may be satisfied by more than one capability or plan.
- The intent carries or references the authority under which work is requested.
- Success criteria are explicit enough to support verification, confidence assessment, and termination.

- Ambiguity is represented as state that can trigger clarification, conservative policy, or human review.

**Example:**

```
intent_id: int-2026-0716-001
objective: summarize the customer incident and recommend next actions
constraints: [confidential-data, us-west-only, max-latency-5s]
authority: support-tier-2
success_criteria: [structured-summary, cited-evidence, confidence>=0.90]
deadline: 2026-07-16T18:00:00Z
```

Intent differs from desired state in a conventional orchestrator. A desired replica count or container image is already close to execution. An intent may require decomposition, discovery, policy interpretation, and negotiation before a concrete execution directive exists. This semantic distance is one reason AOS requires a separate control plane.

An intent may originate from a human, another agent, an enterprise application, a monitoring system, or a policy-triggered automation. The source affects trust and authority but not the core object model.

## 6.3 Context

### Context

Context is the bounded operational state made available to a decision or execution step. It includes relevant conversation state, retrieved knowledge, memory, environment state, prior actions, policy facts, data classification, and temporal information. Context is not identical to memory: memory is a persistence mechanism, whereas context is the selected and scoped input surface for a particular decision.

$$X_t = select(M, K, Env_t, H_t, \Pi, B)$$

Here, M denotes eligible memory sources; K, knowledge sources; Env_t, the current environment state; H_t, the execution history; Π, policy-relevant facts; and B, the context budget governing size, privacy, and relevance.

Key properties:

- Context is scoped to an intent, plan step, actor, and authority domain.
- Every included record should retain provenance, freshness, classification, and access scope.
- Context construction is observable and auditable because omission or inclusion can materially change behavior.
- Context may be different for two actors participating in the same intent because their authority and role differ.
- Context should be treated as an input to confidence and risk, not as inherently correct.

**Example:**

```
context_id: ctx-44
intent_id: int-2026-0716-001
sources:
  - case_record: CASE-12345
  - policy_snapshot: pol-91
  - incident_timeline: trace-abc
redactions: [customer-ssn]
freshness_deadline: 2026-07-16T17:30:00Z
token_budget: 12000
```

Context engineering improves the information surface available to reasoning, but it is not a substitute for control. A well-constructed context can still be used to perform an unauthorized action. AOS therefore treats context selection, policy enforcement, and runtime constraints as complementary responsibilities.

Context versioning is important for reproducibility. If the same plan step is retried with different evidence or memory, the runtime should emit a new context identifier rather than reusing the old one.

## 6.4 Capability

### Capability

A Capability is a stable, discoverable description of what an actor or system can perform. It separates the requested outcome from the implementation that realizes it. A capability may be provided by an agent, model, tool, API, workflow, human, service, or infrastructure function.

$$C = \{id, type, Q, O, V, R, T, L, K, H, Z\}$$

Here, id is the stable capability identifier; type is the capability class; Q is the input/output contract; O is the set of operational obligations; V is the version; R is regional and protocol availability; T is the trust profile; L is the latency profile; K is the cost profile; H is health and reliability evidence; and Z contains security and compliance attributes.

Key properties:

- Capability identity is stable across provider replacement.
- A capability exposes machine-readable input, output, side-effect, and error contracts.
- A capability can have several providers with different trust, cost, latency, locality, and reliability profiles.
- Side effects and destructiveness are part of capability metadata rather than hidden implementation details.
- Capability selection occurs only after hard policy and authority constraints are satisfied.

**Example:**

```
capability_id: ai.case.summary.generate
type: workflow
version: 1.0
inputs: [case_record, evidence]
outputs: [summary, citations, confidence]
side_effects: none
protocols: [mcp, http]
regions: [us-west, us-east]
compliance: [soc2, confidential-data]
observability_profile: full-lineage
```

Capability is deliberately broader than tool. A tool usually describes an invocation endpoint. A capability describes an outcome contract and may be backed by a workflow containing several tools, a human process, or a distributed agent team. This abstraction matches practical enterprise patterns in which developers reuse shared functions without binding to the underlying prompt, model, or workflow.

The planned AOS-0004 document is intended to define discovery, registration, metadata, negotiation, health, and federation in detail. AOS-0001 establishes capability as the stable unit of selection.

## 6.5 Provider

### Provider

A Provider is a concrete implementation that realizes one or more capabilities. The provider owns an endpoint or execution mechanism, a versioned implementation, operational metadata, and evidence about health and behavior.

$$p_i^{(j)} : X_j \rightarrow Y_j \times E_j$$

Provider i’s implementation of capability j is modeled as a function from the authorized input space to an ordered pair containing the result and its execution evidence.

Key properties:

- Providers are replaceable and independently versioned.
- A provider may expose several protocol bindings for the same capability.
- Provider health and trust can change without changing the capability identifier.
- A provider can be local, remote, human-operated, cloud-hosted, edge-hosted, or federated.

**Example:**

```
capability: ai.case.summary.generate
providers:
  - id: internal-workflow-west
    protocol: mcp
    region: us-west
  - id: managed-model-east
    protocol: http
    region: us-east
  - id: human-review-service
    protocol: a2a
    region: global
```

## 6.6 Plan

### Plan

A Plan is a bounded and reviewable decomposition of an intent into steps, dependencies, candidate capabilities, checkpoints, and failure actions. A plan is not authorization to execute; it becomes executable only after the required policy, authority, trust, and confidence decisions are attached.

$$P_n = (V, E, \phi, \omega)$$

Here, V is the set of plan steps; E is the set of dependency edges; φ contains step constraints and preconditions; and ω contains recovery, compensation, escalation, and termination rules.

Key properties:

- Plans are versioned and linked to the originating intent.
- Every step declares required capability classes, inputs, outputs, and side-effect expectations.
- A plan can be deterministic, partially ordered, conditional, or generated incrementally.
- Changes to a plan after authorization may require policy re-evaluation.
- The runtime records actual execution separately from the proposed plan.

**Example:**

```
plan_id: plan-77
intent_id: int-2026-0716-001
steps:
  - retrieve-case-evidence
  - classify-and-summarize
  - verify-citations
  - deliver-result
checkpoints: [after-retrieval, before-delivery]
failure_policy: [retry-once, fallback-provider, human-escalation]
```

## 6.7 Policy

### Policy

Policy is a machine-evaluable rule or rule set that determines eligibility, constraints, obligations, approvals, and response actions. Policy converts organizational, legal, security, privacy, economic, and operational requirements into deterministic control decisions.

$$\pi(I, a, c, x, s) \rightarrow \{allow, deny, modify, approve\} \times \Omega$$

Here, π is the policy-decision function; I is the intent; a is the actor and authority context; c is the candidate capability or provider; x is the current context; and s is the current system state. The function returns a structured decision rather than free-form text: an outcome drawn from allow, deny, modify, or approve, paired with a possibly empty set of obligations Ω, so that compound decisions such as allow-with-obligations are directly expressible. The outcome approve indicates that an additional approval authority is required before execution.

Key properties:

- Policy evaluation is deterministic for a fixed input and policy version.
- A policy decision includes the rule version, result, justification, and obligations.
- Hard policy constraints are evaluated before utility ranking.
- Policy obligations remain attached to delegation and execution.
- Policy can require human approval or stronger runtime isolation for high-impact actions.

**Example:**

```
policy_id: customer-data-egress-v3
if:
  data_classification: confidential
then:
  allowed_regions: [us-west]
  external_network: deny
  required_obligations: [full-lineage, encrypted-storage]
  human_approval_if: side_effecting_action
```

## 6.8 Identity and Authority

### Identity and Authority

Identity establishes who or what an actor is; Authority establishes what that actor may request, approve, delegate, or execute. AOS separates the two because a strongly authenticated actor may still lack permission for a particular intent or capability.

$$A = \{principal, roles, permissions, scope, validity, source\}$$

Here, principal identifies the human, agent, service, device, or organization; roles are contextual roles; permissions are allowed actions; scope bounds data, resources, purpose, and geography; validity captures time and revocation state; and source identifies the policy, contract, or delegation from which authority originates.

Key properties:

- Authority is explicit, scoped, time-bounded, and revocable.
- Authority may originate from a human, organization, policy, contract, or trusted service.
- Proof of identity, proof of human, workload identity, and device attestation are trust inputs; none alone defines authority.
- Consequential execution must be traceable to an authority source.

**Example:**

```
principal: user-42
role: support-tier-2
permissions: [read_case, request_summary, generate_summary]
scope:
  customer: ACME
  region: us-west
valid_until: 2026-07-16T20:00:00Z
source: enterprise-rbac-policy-v12
```

## 6.9 Delegation

### Delegation

Delegation is the controlled transfer of a bounded subset of authority from one actor to another while preserving parentage, constraints, and revocation. Delegation makes authority usable across distributed agents and services without converting every downstream actor into the original principal.

$$d_k = (u, v, S_k, B_k, T_k, parent_k)$$

Here, u is the delegating actor; v is the delegated actor; 𝒮_k is the permitted scope; ℬ_k contains budgets and obligations; 𝒯_k is the validity interval; and parent_k links the delegation to its parent delegation or root authority.

Key properties:

- Child scope cannot exceed parent scope.
- Child expiry cannot exceed parent expiry.
- Child budgets cannot exceed remaining parent budgets.
- Delegation lineage remains observable across protocols and providers.
- Revocation and termination semantics are explicit even when downstream work is asynchronous.

**Example:**

```
delegation_id: del-002
parent_delegation_id: del-001
root_intent_id: int-2026-0716-001
delegator: planner-agent-v3
delegatee: summary-agent-v1
allowed_actions: [read_case, generate_summary]
denied_actions: [modify_case, external_network]
valid_until: 2026-07-16T17:17:10Z
```

The architecture requires delegation-scoped observability because conventional trace structure alone may be insufficient to reconstruct which actions occurred under a particular delegated authority [4].

## 6.10 Trust

### Trust

Trust is an evaluated relationship describing the degree to which an actor, provider, artifact, or environment is acceptable for a particular intent and policy context. Trust is contextual and evidence-based rather than a universal scalar reputation.

$$T_i = f(identity, attestation, provenance, policy, history, behavior, context)$$

The trust function combines verifiable identity, platform or software attestation, artifact provenance, organizational policy, historical reliability, observed behavior, and the current operational context. Trust is evaluated for a specific intent and is not a universal reputation score.

Key properties:

- Trust can differ by capability, data class, region, and requested side effect.
- Trust decisions are versioned and evidence-backed.
- Trust may decay when evidence becomes stale or behavior changes.
- Trust is an input to feasibility, ranking, confidence, and runtime isolation.

**Example:**

```
subject: internal-workflow-west
capability: ai.case.summary.generate
trust_level: high
evidence: [signed-image, workload-attestation, 30-day-slo-history]
restrictions: [confidential-data-only, no-write-side-effects]
valid_until: 2026-07-17T00:00:00Z
```

## 6.11 Confidence

### Confidence

Confidence is a multidimensional operational assurance signal describing how strongly the current evidence supports a proposed plan, action, or result. It is not equivalent to the probability assigned by one model. It is a system-level object that combines uncertainty and evidence from several components.

$$c = [c_{intent}, c_{context}, c_{data}, c_{model}, c_{capability}, c_{policy}, c_{execution}, c_{temporal}]$$

The components represent intent clarity, context adequacy, data quality, model calibration, provider or capability evidence, policy certainty, runtime validation, and temporal freshness. The vector is retained so that one weak component is not hidden by a single aggregate value.

Key properties:

- Confidence retains component evidence and uncertainty rather than collapsing immediately to an opaque score.
- Confidence is evaluated before, during, and after execution.
- Confidence thresholds are policy- and impact-dependent.
- Confidence drives control actions: proceed, proceed with warning, retry, fallback, escalate, or terminate.
- Confidence propagation across steps must state assumptions about dependence among sources.

**Example:**

```
confidence_id: conf-88
components:
  intent: 0.96
  context: 0.88
  data: 0.86
  model: 0.84
  capability: 0.93
  policy: 1.00
execution: 0.91
  temporal: 0.95
decision: proceed_with_warning
reason: model and data evidence below component warning threshold
```

## 6.12 Evidence and Provenance

### Evidence and Provenance

Evidence is the set of observable artifacts used to justify and reconstruct a control or execution decision. Provenance describes the origin, transformation, custody, and dependencies of those artifacts.

$$E_o = \{ telemetry, decisions, attestations, validations, artifacts, lineage \}$$

Here, telemetry includes traces, metrics, logs, and profiles; decisions include policy, trust, and confidence outputs; attestations include identity and environment claims; validations include tests and verifiers; artifacts include inputs and outputs; and lineage records dependency and delegation edges.

Key properties:

- Evidence is immutable or tamper-evident according to risk.
- Evidence references exact versions of policies, models, providers, contexts, and plans.
- Evidence supports both online control and retrospective audit.
- Provenance records transformations and supersession rather than only current values.
- Causal dependencies are represented separately from wall-clock order.

**Example:**

```
event: capability.selected
intent_id: int-2026-0716-001
provider_id: internal-workflow-west
policy_decision: allow-with-obligations
trust_evidence: attestation-44
registry_snapshot: reg-20260716-1712
trace_id: 4bf92f3577b34da6a3ce929d0e0e4736
span_id: 00f067aa0ba902b7
```

## 6.13 Execution

### Execution

Execution is the realization of an authorized intent through one or more selected capabilities under explicit policy, authority, confidence, budget, and observability constraints. It is the actual runtime path, not merely the proposed plan.

$$E_x = execute(D, P_n, C, X, R)$$

Here, D is the authorized execution directive; P_n is the approved plan version; C is the selected set of capabilities and providers; X is the scoped runtime context; and R is the set of runtime resources and constraints.

Key properties:

- Execution has a unique identifier and monotonically recorded lifecycle.
- Actual provider calls and side effects are recorded independently of the plan.
- Execution may branch, retry, compensate, fall back, escalate, or terminate.
- Every consequential action is linked to intent, authority, delegation, and policy.
- Execution emits results and evidence, including residual risk and confidence.

**Example:**

```
execution_id: exec-456
intent_id: int-2026-0716-001
plan_id: plan-77
provider_id: internal-workflow-west
state: verifying
obligations: [full-lineage, no-external-network]
failure_policy: [retry-once, fallback-provider, human-escalation]
```

## 6.14 Semantic Resource Model

Traditional operating systems manage resources such as CPU time, memory, files, sockets, and device access. AOS additionally manages semantic resources that affect whether and how work should proceed. A semantic resource may be scarce, revocable, jurisdiction-bound, or shared across tenants even when it does not correspond to a physical device.

Examples include an authority budget, token budget, context window, model quota, human-approval capacity, data residency allowance, trust relationship, confidence threshold, evidence-retention budget, and time-to-decision. Treating these as managed resources enables scheduling and policy to account for the true operating constraints of agentic systems.

*Table 5. Relationship between conventional and semantic resources.*

| Conventional resource | AOS semantic resource | Operational relationship |
| --- | --- | --- |
| **CPU / accelerator time** | Capability capacity | A capability may consume one or more compute pools but is selected by semantic fit and policy. |
| **Memory** | Context budget and durable memory | The system chooses which state is visible and which state is persisted. |
| **File / object** | Evidence and provenance artifact | Artifacts carry classification, lineage, retention, and authority metadata. |
| **Network path** | Trust and jurisdiction path | Routing considers network delay together with administrative and regulatory boundaries. |
| **Process identity** | Actor and delegation identity | The executing process is linked to the semantic actor and authority chain. |
| **Quota** | Cost, token, and approval budget | AOS allocates non-compute budgets that constrain reasoning and execution. |

## 6.15 Conceptual Invariants

- Every execution originates from an immutable intent or an explicitly identified system objective.
- Every provider invocation realizes a declared capability contract.
- Every consequential action is covered by an authority chain and applicable policy decision.
- Delegated authority is monotonically non-expanding unless an independent authority explicitly grants additional scope.
- Control decisions and execution actions are separately identifiable and jointly traceable.
- Confidence decisions retain evidence and map to explicit runtime actions.
- Audit and observability records preserve version, lineage, and causal relationships sufficient for declared conformance profiles.

These invariants are foundational. Later AOS specifications may define stronger requirements, but they should not weaken the semantic links that make end-to-end governance possible.

# 7. Design Principles

The design principles translate the foundational concepts into architectural rules. They are not implementation recipes; they define the properties that implementations should preserve when making different engineering tradeoffs.

## 7.1 Separation of Control and Execution

AOS distinguishes decisions about what may happen from mechanisms that perform the work. The same process or cluster may host both planes, but the decision record, execution directive, and observed action remain separate objects. This separation supports independent evolution, policy review, fault containment, and multi-runtime interoperability. It also prevents a runtime from treating local success as proof that the action was authorized or appropriate.

## 7.2 Capability-Centric Abstraction

Applications request capabilities instead of binding directly to agents, models, or vendors. The capability contract is stable; providers are replaceable. This allows AOS to route according to policy, trust, cost, latency, locality, reliability, and evidence. It also creates a clean contribution model for a public landscape: projects can register the capabilities they provide without claiming to implement the entire operating architecture.

## 7.3 Authority Preservation and Least Delegation

Authority must remain visible from the root objective to the final side effect. Delegation follows least-authority principles: the child receives only the scope, budget, and lifetime necessary for the delegated work. The architecture rejects implicit authority expansion through nested agent calls, tool adapters, or asynchronous queues.

## 7.4 Deterministic Enforcement Around Probabilistic Reasoning

Probabilistic components may interpret intent, propose plans, rank alternatives, and generate outputs. Actions that access protected data or modify external state should cross deterministic policy, schema, authorization, and runtime enforcement boundaries. This principle is not a demand that the entire system be deterministic. It is a requirement that consequential boundaries be explicit and enforceable.

## 7.5 Confidence as a Control Signal

Confidence is operational only when it changes behavior. AOS therefore associates confidence thresholds with actions and impact classes. A low-confidence classification may request more evidence; a low-confidence financial transfer must not proceed; an uncertain but reversible summary may proceed with warning. The mapping from evidence to action is policy-governed and auditable.

## 7.6 Observability by Construction

Semantic observability fields are created as the system operates rather than reconstructed later from log text. Intent identifiers, capability and provider identifiers, policy versions, trust evidence, context references, delegation lineage, confidence records, and causal links propagate with execution. Existing standards such as W3C Trace Context and OpenTelemetry can carry or reference this data [9, 10].

## 7.7 Explicit State and Legal Transitions

Long-running agentic systems require declared lifecycle states. Intents, plans, delegations, capabilities, executions, approvals, and confidence assessments each have legal transitions. Explicit state reduces drift, enables idempotent recovery, and provides a foundation for testing. The runtime may add internal states but cannot silently rewrite externally visible history.

## 7.8 Runtime, Model, Cloud, and Operating-System Agnosticism

AOS defines semantic responsibilities and contracts, not exclusive implementations. A capability may be invoked through MCP, A2A, HTTP, a queue, a local function, or a future protocol. Execution may run on Linux, Windows, containers, serverless infrastructure, edge devices, or specialized accelerators. Agnosticism is constrained by declared profiles: an implementation must state what it supports and what remains external.

## 7.9 Federation over Global Monoliths

Global operation does not imply one global controller. Organizations and regions may retain local policy, trust, data, and runtime control while sharing capability advertisements and interoperable evidence. Federation reduces concentration and supports sovereignty, but it introduces stale state, policy conflicts, and revocation challenges. These tradeoffs must be explicit.

## 7.10 Reliability as an End-to-End Property

Model accuracy, runtime availability, data freshness, policy correctness, infrastructure health, and human operations jointly determine reliability. AOS assigns responsibility across planes and preserves the evidence needed to locate failures. No single confidence score or uptime metric is sufficient.

## 7.11 Human Governability and Exception Handling

Humans remain responsible for policy, approval thresholds, exceptions, appeals, and accountability. Human-in-the-loop is not one generic step; it includes approval before action, supervisory review, emergency stop, post-action review, and dispute resolution. The architecture should make these roles explicit rather than hiding them in application UI.

## 7.12 Economic and Physical Awareness

AOS decisions account for latency, queueing, bandwidth, energy, cost, carbon intensity, accelerator topology, and thermal or capacity limits. Intelligence is not free of physics. A geographically remote provider has a propagation-delay lower bound; a model with low average latency may have unacceptable tail latency; a cheaper provider may increase risk or failure cost. Scheduling therefore uses multi-objective constraints rather than model quality alone.

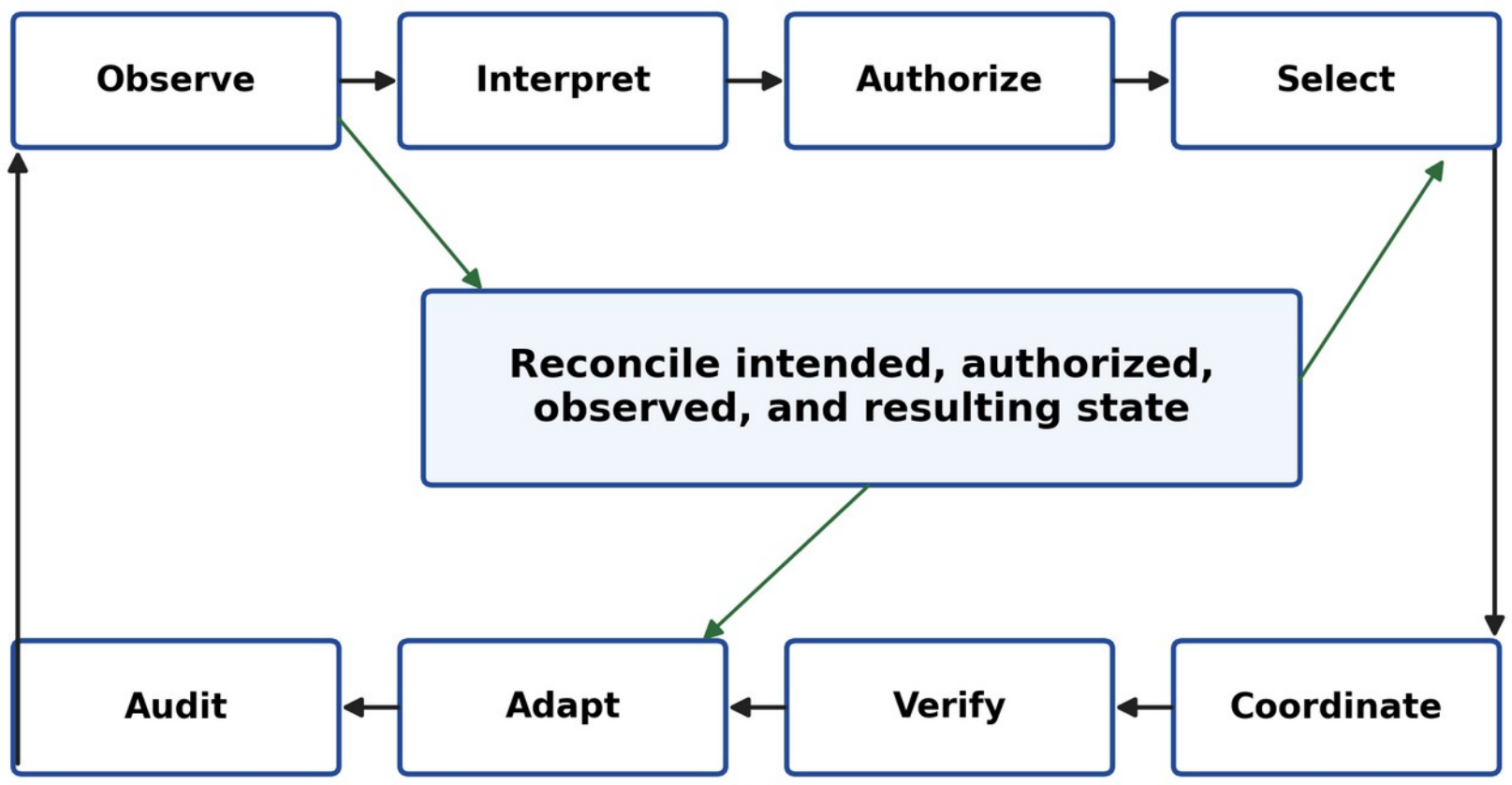

*Evidence and policy close the loop; every transition remains observable and auditable.*

*Figure 3. Evidence-driven AOS control loop. The architecture reconciles intended, authorized, observed, and resulting state through an auditable closed-loop process.*

## 7.13 Principle Interactions and Tensions

The principles can conflict. Strong observability may increase privacy exposure. Strict deterministic boundaries may increase latency. Global capability discovery may conflict with sovereign control. Runtime agnosticism may reduce opportunities for deep optimization. AOS does not eliminate these tensions; it makes them visible and provides the objects through which policy can resolve them.

A conforming design should document which principle dominates when tradeoffs cannot be simultaneously satisfied. For example, an emergency medical system may prefer availability and human escalation over cost optimization, while a sovereign government deployment may prefer locality and policy autonomy over global provider diversity.

# 8. Reference Architecture

## 8.1 Architectural Overview

AOS is defined by a boundary containing two internal planes. Plane 1, Control & Governance, interprets objectives and creates authorized, evidence-bearing decisions. Plane 2, Runtime & Coordination, translates those decisions into coordinated work across agents, workflows, tools, models, memory, and infrastructure. The planes exchange explicit directives, state, events, feedback, and telemetry.

Below the AOS boundary are external platform services, conventional operating systems, and physical or virtual infrastructure. They remain essential and may supply major portions of the implementation. The boundary is not a claim that AOS owns every service it manages. It is a declaration of architectural responsibility: AOS must know which external capability is being used, under what constraints, and with what evidence.

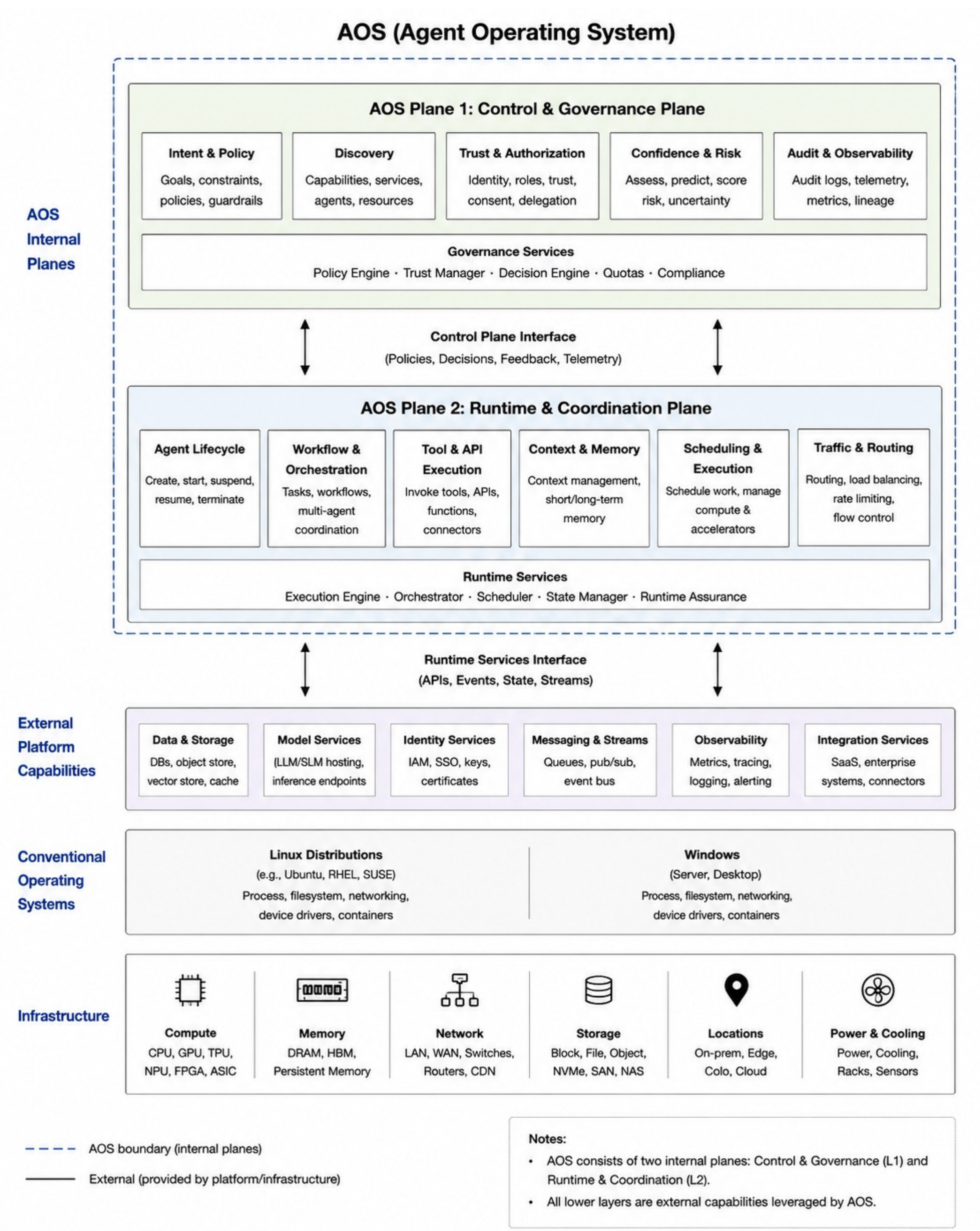

*Figure 4. AOS unified reference architecture. Only the Control & Governance Plane and Runtime & Coordination Plane are inside the AOS boundary; platform services, conventional operating systems, and infrastructure remain external managed dependencies.*

## 8.2 Plane Boundary

The AOS boundary is semantic. Plane 1 owns authoritative intent, policy decisions, trust evaluations, delegation grants, confidence actions, audit obligations, and human-control requirements. Plane 2 owns active execution state, provider bindings, runtime queues, workflow progress, context delivery, scheduling decisions, traffic state, and assurance responses. A control component may request execution, but it does not directly claim that the work succeeded. A runtime component may report success, but it does not retroactively authorize the action.

The boundary can be implemented with process APIs, event streams, in-memory contracts, or a distributed control fabric. What matters is that decision and execution state can be identified and correlated. This enables testing, replay, forensic analysis, and independent replacement of runtime components.

## 8.3 Control & Governance Plane Summary

*Table 6. Control & Governance Plane responsibilities.*

| Responsibility | Primary output | Representative failure |
|---|---|---|
| **Intent management** | Normalized IntentEnvelope | Ambiguous, conflicting, or unauthenticated objective |
| **Capability discovery** | Feasible provider set and ranking evidence | Stale registry, missing metadata, false advertisement |
| **Policy and trust** | Allow, deny, modify, approve, and obligations | Conflicting policy, stale attestation, trust misclassification |
| **Confidence and risk** | ConfidenceDecision and required action | Miscalibration, dependence error, stale evidence |
| **Delegation governance** | Bounded DelegationContext | Scope expansion, missing revocation, broken lineage |
| **Audit and observability** | Semantic events and evidence requirements | Dropped context, tampering, incompatible schemas |
| **Human oversight** | Approval, exception, escalation, or stop decision | Unavailable approver, unclear accountability |

## 8.4 Runtime & Coordination Plane Summary

*Table 7. Runtime & Coordination Plane responsibilities.*

| Responsibility | Primary state | Representative failure |
|---|---|---|
| **Agent lifecycle** | Created, active, paused, migrated, revoked, terminated | Orphaned agent, duplicate execution, failed revocation |
| **Workflow coordination** | Plan-step and dependency state | Loop, illegal transition, compensation failure |
| **Model and tool routing** | Provider binding and request state | Timeout, protocol mismatch, side-effect surprise |
| **Context and memory coordination** | Versioned context and memory references | Stale context, leakage, contradiction, contamination |
| **Scheduling** | Queues, budgets, placement, concurrency | Head-of-line blocking, starvation, oversubscription |
| **Traffic management** | Routes, retries, rate limits, backpressure | Retry storm, overload, regional black hole |
| **Runtime assurance** | Checks, circuit breakers, isolation, cancellation | Delayed detection, incomplete containment |

## 8.5 External Platform Services

Platform services include model-serving endpoints, retrieval and vector systems, data stores, knowledge graphs, identity providers, secrets, policy decision points such as Open Policy Agent [22], message brokers, observability backends, API gateways, and enterprise connectors. AOS may consume these services through capability and provider contracts. A service remains external even when it is tightly integrated with the runtime.

This distinction prevents architectural scope from expanding to “everything AI.” A vector database is not part of AOS merely because context depends on it. A policy engine may implement a control-plane interface without becoming the whole control plane. The paper distinguishes managed dependencies from internal plane responsibilities.

## 8.6 Conventional Operating Systems and Hardware

Linux, Windows, hypervisors, and container runtimes manage local processes, filesystems, memory, sockets, credentials, devices, and isolation. AOS uses adapters and enforcement translators to connect semantic directives to these mechanisms. The adapter may map an execution budget to cgroups, an allowed action set to a sandbox profile, a network policy to firewall or service-mesh rules, or a delegated identity to workload credentials.

Hardware and facilities constrain what the runtime can achieve. Accelerator type, memory capacity, interconnect topology, storage latency, network distance, clock quality, power budget, and thermal limits are inputs to scheduling and confidence [19]. AOS does not directly control every device; it receives abstractions and telemetry from platform and infrastructure systems.

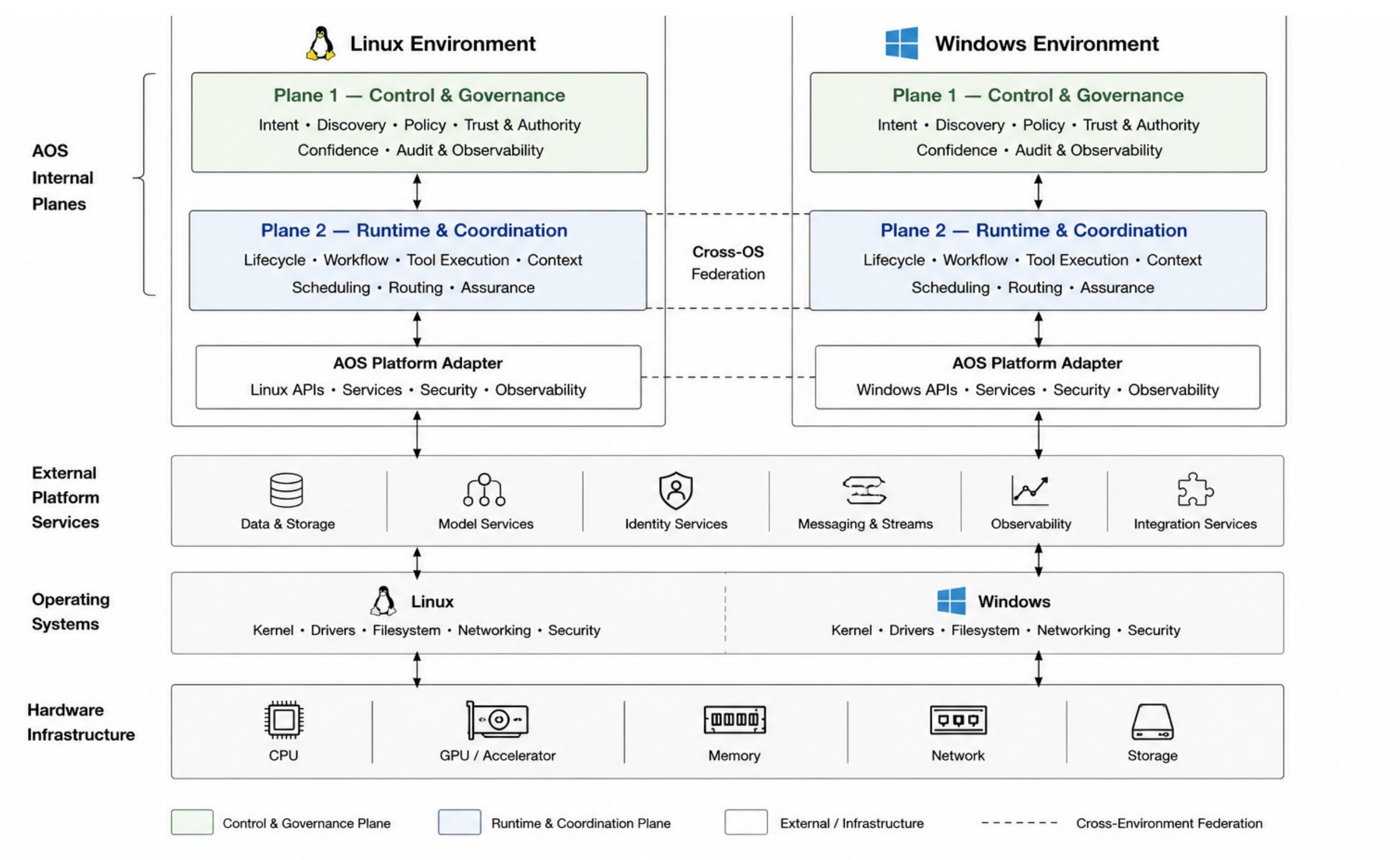

*Figure 5. Identical AOS planes operate over Linux and Windows through platform adapters while sharing external services and hardware resources.*

## 8.7 State, Feedback, and Reconciliation

The architecture maintains four related forms of state: intended state, authorized state, observed state, and resulting state. Intended state originates in the intent and success criteria. Authorized state includes policy decisions, selected capability, delegated scope, budgets, and obligations. Observed state is obtained from runtime and platform telemetry. Resulting state records actual outcomes and residual risk.

AOS controllers reconcile these states. If observed state diverges from authorized state, the system may cancel, isolate, or escalate. If resulting state fails success criteria, it may retry or select a fallback. If the intent changes, a new version enters the lifecycle rather than mutating the audit history. Reconciliation is therefore semantic and operational, not only resource-based.

## 8.8 Deployment Independence

Logical planes do not imply mandatory network hops. An embedded implementation may call control and runtime modules in memory. A centralized deployment may expose them as services. A federated deployment may exchange signed capability advertisements, policy claims, and execution evidence across organizations. The same canonical objects can be serialized differently according to profile.

This property is important for research and adoption. The architecture can be prototyped in aos-core without requiring a hyperscale infrastructure, while the same boundaries remain meaningful in a multi-region enterprise deployment.

## 8.9 Conformance at the Architecture Level

*Table 8. Architecture-level conformance terminology.*

| Level | Meaning |
|---|---|
| **AOS-aware** | Consumes or emits selected AOS objects or semantic events but implements a narrow function. |
| **AOS-compatible** | Implements one or more declared plane interfaces without violating core invariants. |
| **AOS-conformant profile** | Implements all mandatory responsibilities and passes tests for a named profile. |
| **AOS reference implementation** | A non-exclusive implementation intended to validate architecture and specifications. |

# 9. Control & Governance Plane

## 9.1 Role and Scope

The Control & Governance Plane is the semantic authority of AOS. It converts an objective into an authorized execution directive. It owns the interpretation of intent, the evaluation of authority and policy, the discovery and ranking of feasible capabilities, the formation of delegation boundaries, the definition of confidence actions, and the evidence obligations that make decisions explainable and reviewable.

The control plane does not execute model inference, run workflow steps, maintain accelerator queues, or manage operating-system resources directly. It may call services that perform analysis, but the authoritative output is a decision record rather than an external side effect. This scope preserves separation from the Runtime & Coordination Plane.

## 9.2 Intent Ingestion and Normalization

Intent ingestion validates the actor, authority source, objective, constraints, context references, success criteria, and temporal requirements. Natural-language requests may require semantic normalization; machine-generated intents may already be structured. The result is an IntentEnvelope that remains stable through the lifecycle.

Normalization should not silently invent authority or remove constraints. If the objective is ambiguous, the plane can request clarification, choose a conservative interpretation, or route to a human. The normalized representation records which information came from the actor, which was inferred, and which policy defaults were applied.

## 9.3 Capability Discovery and Feasibility

The control plane queries local or federated registries for capabilities matching the intent. Discovery returns provider metadata, not merely endpoint addresses. The control plane evaluates input and output compatibility, side effects, supported protocols, provider version, trust evidence, regions, data handling, latency and cost profiles, health, quota, and observability support.

Feasibility precedes ranking. A provider that is unauthorized, outside the allowed jurisdiction, incompatible with the data classification, or unable to satisfy a required audit profile is excluded regardless of its predicted quality or cost. This ordering prevents optimization from overriding governance.

## 9.4 Policy Evaluation and Obligations

Policy evaluation combines the intent, actor, capability, provider, context, risk class, and current system state. Decisions may allow, deny, modify, require approval, or attach obligations. Examples of obligations include full-lineage telemetry, no external network, encrypted temporary storage, a specific region, bounded token cost, a post-execution verifier, or human approval before a side effect.

Policy composition can be difficult when rules originate from organizations, regulators, providers, and users. The architecture requires the final decision to identify the applicable policy versions and the conflict-resolution rule. A later specification should define policy federation and precedence in more detail.

## 9.5 Trust and Identity Evaluation

Trust evaluation binds identity evidence, provider provenance, platform attestation, organizational relationship, historical behavior, and capability-specific reliability. Trust is evaluated relative to the requested action. A provider may be trusted to summarize public documents but not to process regulated customer data or perform financial transactions.

The control plane can require stronger runtime isolation when trust is lower, but it should not use isolation as a substitute for authorization. A technically contained but unauthorized action remains unauthorized.

## 9.6 Confidence and Risk Decision

The control plane defines confidence requirements before execution and evaluates evidence after planning or verification. A high-impact action may require high confidence, independent verification, and human approval. A reversible or informational action may proceed with warning. The decision includes the threshold, component evidence, uncertainty, rationale, and required control action.

Risk combines probability, impact, and exposure. Confidence and risk are related but not interchangeable. High confidence that a harmful action will succeed increases risk; low confidence in a harmless draft may be acceptable. The architecture therefore treats confidence as an evidence variable and risk as a policy-relevant decision variable.

## 9.7 Delegation Governance

When work is delegated, the control plane creates a DelegationContext that references the root intent and parent authority. The context contains the delegated actor, allowed capabilities, data and resource scope, time interval, budgets, obligations, confidence thresholds, and revocation information. Downstream delegation can only narrow this context unless an independent authority participates.

The control plane records delegation as a first-class event. This allows the system to answer who initiated the work, which authority was transferred, what downstream actions occurred, which data was accessed, and whether the scope changed. The requirement is stronger than ordinary request tracing because delegation may span asynchronous and heterogeneous systems.

**Authority and Delegation Chain**

*Figure 6. Authority and delegation chain from human objective to kernel or system execution, with scope and lineage preserved at every transition.*

## 9.8 Audit, Provenance, and Observability Obligations

The control plane determines what evidence the runtime must emit. Low-risk operations may require standard traces and results. High-risk operations may require complete input and output hashes, policy versions, context references, delegation records, verifier results, resource usage, and tamper-evident storage.

Audit is not identical to logs. Logs are implementation outputs; an audit record is a governed evidence object with retention, access, integrity, and accountability requirements. AOS can use OpenTelemetry and W3C Trace Context as transport and correlation substrates while extending them with semantic references [9, 10].

## 9.9 Human Oversight and Control

Human oversight occurs at several points: clarification of ambiguous intent, approval before execution, review of high-risk provider selection, intervention during execution, review of uncertain results, exception handling, and appeal. The control plane routes each request to an accountable role rather than a generic “human.”

A human decision is itself evidence. The record includes the approver identity, authority, information presented, decision, time, and any modifications. Human unavailability is a failure mode that policy must handle explicitly through timeout, alternative approver, safe fallback, or termination.

## 9.10 Authorized Execution Directive

The principal output of the Control & Governance Plane is an authorized execution directive containing:

- intent and plan identifiers;
- selected capability and provider, or a bounded candidate set;
- policy and trust decisions with versions and rationale;
- delegation context and authority scope;
- confidence thresholds and verification requirements;
- runtime placement, isolation, budget, and protocol constraints;
- observability, audit, provenance, and retention obligations;
- failure actions such as retry, fallback, escalation, compensation, or termination.

**Plane 1 invariant.** The Control & Governance Plane authorizes and constrains work; it MUST NOT assert that the work has occurred or succeeded. Runtime evidence is REQUIRED to close the loop.

# 10. Runtime & Coordination Plane

## 10.1 Role and Scope

The Runtime & Coordination Plane is the operational machinery of AOS. It receives authorized execution directives and coordinates their realization across providers, frameworks, tools, models, humans, data services, operating systems, and infrastructure. It owns active execution state and emits evidence that allows the control plane to verify the result.

The runtime plane should integrate existing engines rather than reimplement them without reason. A LangGraph workflow, a conventional BPM engine, an MCP server, an A2A agent, a model gateway, a human task service, or a custom Python function can all serve as providers. The runtime provides common lifecycle, routing, scheduling, state, assurance, and protocol boundaries around them.

## 10.2 Agent Lifecycle Management

Agent lifecycle includes create, initialize, activate, pause, resume, quiesce, migrate, revoke, terminate, and garbage-collect. The managed object may be a persistent agent identity or a transient runtime instance. AOS distinguishes identity from instance so that an agent can migrate or restart without losing authority and audit continuity.

Lifecycle transitions are policy- and state-constrained. Revocation must stop new delegated work and define how in-flight operations are handled. Migration must preserve context, delegation, checkpoints, and trace continuity. Termination must emit final state and evidence rather than simply deleting the process.

## 10.3 Workflow and State Coordination

Workflow coordination executes plans and legal state transitions. It manages dependencies, checkpoints, retries, compensation, timeouts, human tasks, and partial results. The runtime distinguishes plan state from actual execution state so that deviations and dynamic replanning remain observable.

AOS does not require a single workflow representation. A finite-state machine, DAG, event-driven process, rule engine, or agent conversation can all be mapped to the canonical execution lifecycle. The adapter is responsible for reporting state and causal relationships in AOS terms.

## 10.4 Model, Tool, API, and Human Routing

Routing binds a capability to a provider and protocol. The runtime applies credentials, delegation context, input schema, rate limits, budgets, timeouts, and observability requirements. It records the actual endpoint and version because the control-plane selection may have allowed several equivalent providers.

Human providers are routed similarly but with different latency and interaction semantics. The runtime can create a review task, deliver the required evidence, wait for a decision, and continue or terminate. Treating humans as capability providers avoids embedding approval logic into every workflow while preserving accountability.

## 10.5 Context and Memory Coordination

Context coordination builds the bounded input for each runtime step. It retrieves authorized records, applies redaction and purpose limitation, records provenance, enforces size and token budgets, and delivers a versioned context reference. The runtime should not assume that passing the entire history is safe or efficient.

Memory coordination manages access to episodic, semantic, procedural, and artifact memory. It tracks freshness, contradiction, supersession, ownership, and retention. When memory changes during a long-running intent, the runtime records whether the plan used the old or new version and may request control-plane re-evaluation.

## 10.6 Scheduling

Runtime scheduling allocates queues, concurrency, token budgets, model capacity, accelerators, regions, human review slots, and deadlines. Agentic workloads create service-time variance: one request may produce a short classification while another generates thousands of tokens or invokes several tools. Request-level FIFO scheduling can therefore create head-of-line blocking. Token-, step-, deadline-, or priority-aware scheduling may reduce latency and improve fairness.

Scheduling is constrained by policy and capability contracts. A faster GPU in another region may be infeasible because of data residency. A lower-cost model may be rejected because the action requires a higher assurance class. The runtime reports dynamic capacity and queue state to the control plane, which can re-evaluate placement and provider selection.

## 10.7 Traffic Management and Backpressure

Traffic management controls routes, load balancing, retries, rate limits, circuit breakers, backpressure, and cross-region fallback. Agentic call graphs can amplify load because one intent may spawn several agents and tools. Unbounded retry and delegation can produce storms. The runtime therefore enforces fan-out, depth, token, time, and cost budgets.

Traffic policies distinguish idempotent and non-idempotent capabilities. A read-only retrieval may be retried automatically; a payment or configuration change requires stronger deduplication and compensation. The provider contract declares idempotency, side effects, and safe retry behavior.

## 10.8 Runtime Assurance

Runtime assurance observes whether execution remains within the authorized envelope. It checks schema, side effects, resource usage, network destinations, process behavior, confidence, and health. Responses include allow, warn, throttle, isolate, cancel, fallback, escalate, or terminate. These responses refine the canonical control actions of Table 13 with enforcement-level mechanisms rather than defining an independent decision vocabulary.

Assurance can operate at several levels. An application validator checks output structure. A gateway restricts tool calls. A sandbox limits filesystem and network access. A container runtime enforces namespaces and cgroups. A kernel mechanism filters syscalls. AOS preserves the relation among these controls and the original authority chain rather than treating them as isolated security events.

## 10.9 Protocol Mediation

Protocol mediation maps canonical AOS objects to external protocols. An MCP adapter may convert an execution directive into a tool invocation with authorization and trace context. An A2A adapter may create an agent task and propagate delegation. An HTTP adapter may apply OAuth credentials and idempotency keys. An event adapter may serialize state changes to a broker using standard event envelopes such as CloudEvents [50].

The adapter does not own policy. It must reject requests whose mandatory constraints or evidence requirements it cannot represent. For example, if a target protocol cannot propagate delegation context required by the conformance profile, the runtime must use another binding or declare the limitation.

## 10.10 Runtime Evidence and Result Handling

The runtime emits execution results together with evidence: actual provider, version, timing, resource usage, context identifiers, tool calls, side effects, validation output, confidence components, errors, and residual risk. The result is not complete until required evidence has been persisted or acknowledged according to policy.

The control plane uses the result to decide whether success criteria are met. A technically successful call can still fail the intent if citations are missing, confidence is insufficient, the wrong data version was used, or an unauthorized side effect occurred. This distinction prevents runtime return codes from becoming the only definition of success.

> **Plane 2 invariant.** The Runtime & Coordination Plane executes and observes authorized work. It MUST NOT expand authority, silently discard obligations, or redefine success criteria without returning to the Control & Governance Plane.

# 11. Interfaces

## 11.1 Interface Philosophy

A reference architecture becomes useful when its boundaries can be implemented by independently evolving components. AOS therefore defines semantic interfaces rather than one mandatory wire protocol. The interface layer specifies which objects cross a boundary, which state is authoritative, which obligations must be preserved, and which failure conditions must be reported. Protocol bindings decide how these semantics are serialized and transported.

The architecture follows three rules. First, providers should depend on contracts rather than concrete control-plane implementations. Second, every request that can produce an external effect must carry or reference intent, authority, policy, delegation, and trace context. Third, an adapter must fail closed when it cannot represent a mandatory constraint or evidence requirement. Silent semantic loss is a compatibility failure, not a successful integration.

## 11.2 Canonical Objects

*Table 9. Canonical AOS interface objects.*

| Object | Producer | Consumer | Purpose |
|---|---|---|---|
| **IntentEnvelope** | Client, human, upstream agent, monitoring system | Control & Governance Plane | Objective, constraints, authority, context references, success criteria, time requirements |
| **CapabilityQuery** | Control & Governance Plane | Registry or providers | Semantic requirements, hard constraints, preferred objectives, protocol and evidence requirements |
| **CapabilityManifest** | Provider or registry curator | Registry and control plane | Contract, provider metadata, side effects, trust, health, region, cost, latency, compliance |
| **PolicyDecision** | Policy evaluator | Control and runtime planes | Allow, deny, modify, approve, obligations, rule versions, rationale |
| **TrustDecision** | Trust evaluator | Control plane | Subject, evidence, contextual trust level, restrictions, expiry |
| **DelegationContext** | Control plane (children may only narrow) | Downstream actor | Root intent, parent delegation, actor, scope, budgets, obligations, expiry, revocation |
| **ExecutionDirective** | Control & Governance Plane | Runtime & Coordination Plane | Selected capability/provider, constraints, credentials, confidence thresholds, failure policy |
| **ExecutionEvent** | Runtime components | Control, observability, audit | Lifecycle transition, actual provider, timestamps, causal links, resource and side-effect data |
| **ConfidenceRecord** | Evaluator or verifier | Control plane and runtime | Component evidence, uncertainty, score or interval, action, rationale |
| **ExecutionResult** | Runtime plane | Control plane and caller | Outcome, artifacts, confidence, evidence, residual risk, completion state |
| **AuditEvent** | All planes | Audit sink | Tamper-evident decision or action record with lineage and retention metadata |

## 11.3 IntentEnvelope

The IntentEnvelope is the root object for user- or system-initiated work. It is designed to preserve the original objective while allowing normalized fields and references to evolve around it.

```
{
  "intent_id": "int-2026-0716-001",
  "actor": {"type": "human", "id": "user-42"},
  "objective": "Summarize the incident and recommend next actions",
  "constraints": {
```

```
    "data_classification": "confidential",
    "allowed_regions": ["us-west"],
    "max_cost_usd": 2.00,
    "max_latency_ms": 5000,
    "deadline": "2026-07-16T18:00:00Z"
  },
  "success_criteria": ["cited_summary", "confidence >= 0.90"],
  "authority_ref": "auth-771",
  "context_refs": ["CASE-12345", "policy-snapshot-91"],
  "trace_id": "4bf92f3577b34da6a3ce929d0e0e4736"
}
```

The envelope may contain natural-language and structured representations simultaneously. The structured representation is authoritative for machine evaluation; the original language is retained as evidence. If the system infers a constraint or default, it records the source and confidence of that inference.

Intent identifiers should be globally unique within the declared federation profile. Implementations may use UUIDs, content-addressed identifiers, or organizational formats, but collision and replay behavior must be documented.

## 11.4 CapabilityManifest and Registry Interface

```
capability:
  id: ai.case.summary.generate
  type: workflow
  version: "1.0"
  description: Generate a cited incident summary
  inputs:
    - name: case_record
      schema: urn:aos:schema:case-record:1
  outputs:
    - name: summary
      schema: urn:aos:schema:cited-summary:1
  side_effects: none
  protocols: [mcp, http]
provider:
  id: internal-workflow-west
  region: us-west
  trust_profile: trust-high
  latency_profile: latency-summary-west-v2
  cost_profile: cost-medium
  compliance: [soc2, confidential-data]
  observability_profile: full-lineage
  health_endpoint: https://registry.example/health/provider-west
```

The registry interface supports registration, update, revocation, lookup, subscription, health, and federation. Registries may be centralized, distributed, or domain-local. A capability advertisement is not a trust guarantee. The control plane validates signatures, provenance, freshness, and policy before use.

Manifest versioning separates capability contract version from provider implementation version. A provider can upgrade its model or workflow without changing the capability version when input, output, side-effect, and semantic behavior remain compatible. Breaking changes require a new contract version.

## 11.5 ExecutionDirective

```
{
  "execution_id": "exec-456",
  "intent_id": "int-2026-0716-001",
  "plan_id": "plan-77",
  "capability_id": "ai.case.summary.generate",
  "provider_id": "internal-workflow-west",
  "policy_decision_ref": "poldec-112",
  "delegation_context_ref": "del-002",
  "evidence_obligations": ["context-lineage", "output-hash"],
  "runtime_constraints": {
    "region": "us-west",
    "timeout_ms": 3500,
    "token_budget": 12000,
    "max_fanout": 3
```

```
  },
  "confidence_policy": {"required": 0.90, "component_warn_below": 0.90, "on_low": ["fallback_provider",
"human_escalation"]},
    "component_warn_below": 0.90,
  "failure_policy": ["retry_once", "fallback_provider", "human_escalation"]
}
```

The directive is a bounded contract between the planes. The runtime may optimize within the declared freedom, such as choosing among pre-approved replicas, but it cannot relax obligations or expand authority. If runtime conditions make the directive infeasible, it returns a replan or policy-recheck event rather than improvising outside the envelope.

Directives should be idempotent or carry an idempotency key. The profile defines whether resubmission creates a new execution, resumes an existing one, or is rejected.

## 11.6 Service Provider Interfaces

AOS implementations should expose runtime-checkable extension contracts analogous to service-provider interfaces. Representative interfaces include:

```
interface CapabilityProvider
interface CapabilityRegistry
interface IntentNormalizer
interface PolicyEvaluator
interface TrustEvaluator
interface ConfidenceEvaluator
interface DelegationContextProvider
interface ExecutionAdapter
interface RuntimeScheduler
interface TrafficManager
interface ContextProvider
interface MemoryProvider
interface AuditSink
interface ObservabilitySink
interface HumanApprovalProvider
```

Each interface defines inputs, outputs, error classes, versioning, cancellation, and observability obligations. A provider can implement several interfaces, but implementations should avoid hidden coupling. For example, a model gateway may implement ExecutionAdapter and ConfidenceEvaluator; the control plane should still be able to replace either function independently.

The SPI model also supports community participation. Vendors and open-source projects can contribute adapters without claiming ownership of the architecture. Conformance tests can validate behavior at the contract boundary.

## 11.7 Semantic Event Model

AOS events use stable semantic names independent of logging backend. A minimum event catalog includes:

*Table 10. Core semantic events.*

| Event | Required correlation | Representative payload |
|---|---|---|
| **intent.received** | intent_id, actor, trace_id | objective hash, authority reference, source |
| **intent.normalized** | intent_id, version, trace_id | structured fields, inferred defaults, ambiguity |
| **capability.discovery.started** | intent_id, query_id | requirements and registry scope |
| **capability.selected** | intent_id, capability_id, provider_id | feasibility evidence, utility score, rejected alternatives |
| **policy.evaluated** | intent_id, decision_id | result, policy versions, obligations, rationale |
| **trust.evaluated** | subject_id, capability_id | evidence, level, restrictions, expiry |
| **delegation.created** | delegation_id, parent_id, root_intent_id | delegator, delegatee, scope, validity |
| **execution.started** | execution_id, provider_id, delegation_id | directive version, runtime placement |
| **confidence.evaluated** | execution_id, confidence_id | components, threshold, action, evidence |

| Event | Required correlation | Representative payload |
| --- | --- | --- |
| **execution.completed** | execution_id, intent_id | result, artifacts, residual risk |
| **execution.failed** | execution_id, intent_id | failure class, partial effects, recovery action |
| **audit.recorded** | event_id, lineage_id | retention class, integrity proof, sink acknowledgment |

## 11.8 Trace, Delegation, and Causal Context

AOS reuses standard trace identifiers and baggage where possible. The W3C Trace Context traceparent field links requests across vendors, while tracestate allows vendor-specific data [10]. AOS semantic context should not place sensitive data directly into propagation headers. Instead, it carries compact identifiers that reference protected evidence stores.

Trace parentage and delegation parentage are different graphs. One delegation can produce several traces, and one trace may contain work under several delegations if a runtime multiplexes tasks. The event model therefore includes both trace_id/span_id and delegation_id/parent_delegation_id. Causal links can additionally represent message or state dependencies that are not captured by synchronous call structure.

**Agentic Trace and Delegation Graph**

*A distributed trace enriched with intent, delegation, policy, confidence, capability, context, and causal edges*

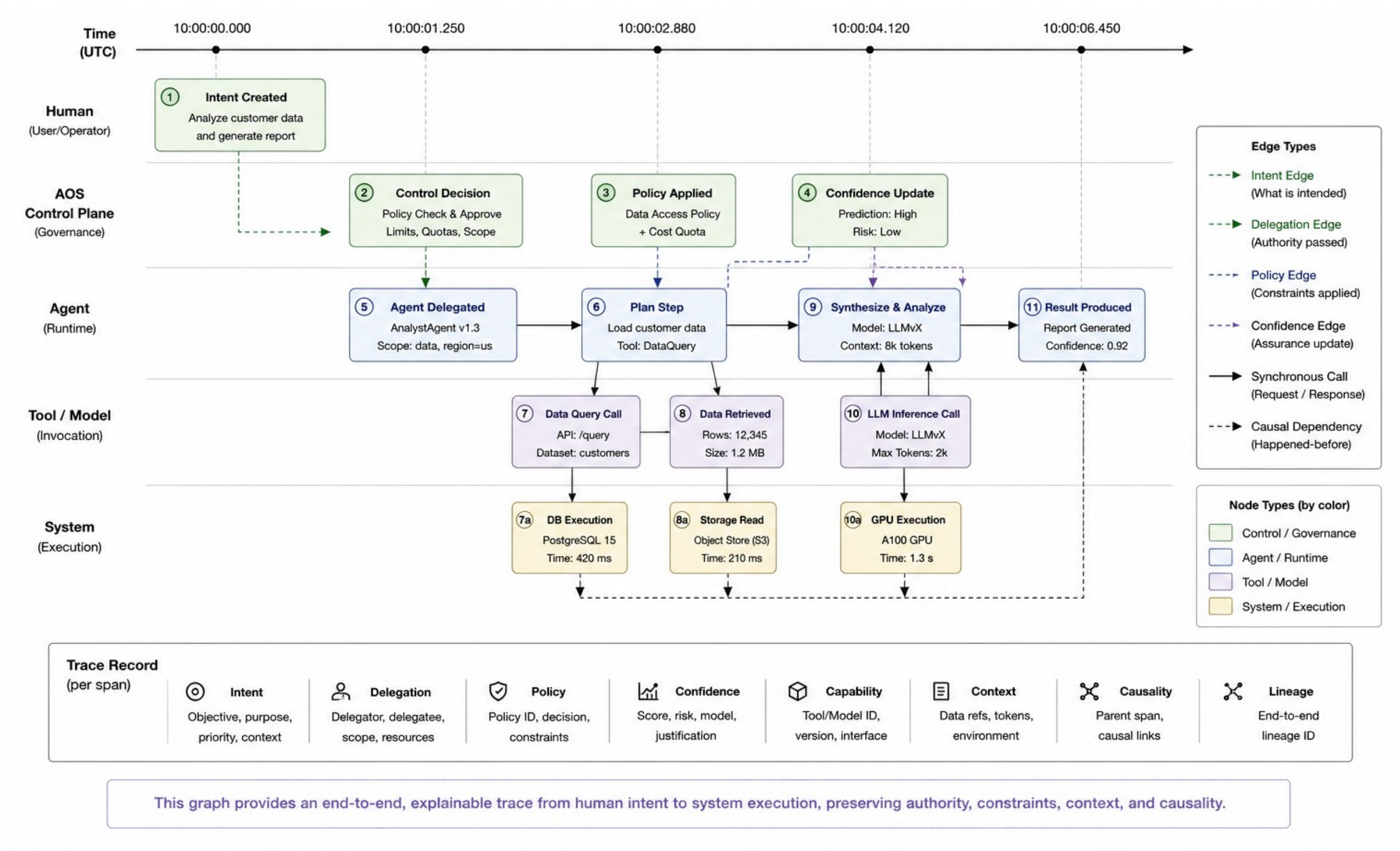

*Figure 7. Agentic trace and delegation graph enriched with intent, policy, confidence, capability, context, causal, and lineage edges.*

## 11.9 Protocol Bindings

*Table 11. Example protocol bindings.*

| Binding | Typical AOS use | Semantic caution |
| --- | --- | --- |
| **MCP** | Tool, resource, prompt, and model-context access | Transport authorization does not replace intent, provider trust, or enterprise policy. |
| **A2A** | Remote agent task exchange and collaboration | Agent task state does not automatically preserve root authority and delegation lineage. |

| Binding | Typical AOS use | Semantic caution |
| --- | --- | --- |
| **HTTP / JSON** | General service and model APIs | Schemas, idempotency, cancellation, and evidence must be profiled explicitly. |
| **Event streams** | Asynchronous state, telemetry, registry, and workflow events | Ordering, replay, duplicate delivery, and revocation lag must be handled. |
| **Local function / library** | Embedded or low-latency providers | In-process calls still require semantic audit and policy boundaries. |
| **Human task protocol** | Approval, review, exception, and escalation | Identity, presentation context, timeout, and accountability are mandatory. |

## 11.10 Versioning and Compatibility

AOS distinguishes specification version, object schema version, capability contract version, provider version, policy version, and execution version. Compatibility claims must name the relevant version. Support for Capability v1.2 does not imply conformance with every AOS profile.

Protocol version negotiation follows the external protocol. MCP, for example, uses date-based version identifiers and negotiates a session version during initialization [7]. An AOS adapter records the negotiated version and whether mandatory semantics were preserved. If a required field cannot be transmitted, the adapter fails or uses an agreed reference mechanism.

## 11.11 Failure Contracts

Every interface declares failure classes rather than returning only generic errors. The classes include invalid intent, unauthorized, policy denied, no feasible capability, provider unavailable, protocol incompatible, context unavailable, deadline exceeded, confidence insufficient, partial side effect, evidence failure, and revocation. A failure record identifies whether retry is safe, whether compensation is required, and whether human review is mandatory.

Failure semantics are essential for closed-loop control. Retrying an unauthorized request is incorrect; retrying a transient network read may be appropriate; repeating a non-idempotent external effect may be dangerous. The runtime uses the contract and directive to select the permitted response.

# 12. Mathematical Model

## 12.1 Purpose and Scope

The mathematical model makes architectural assumptions explicit. It is not intended to prescribe one optimization algorithm. It defines the objects, constraints, and objective classes that a conforming implementation may instantiate. Worked examples are illustrative and do not represent benchmark results. Several equations are decomposition or accounting identities rather than analytical results; they fix shared vocabulary and cost structure for later specifications and conforming implementations.

Let an AOS domain contain actors, intents, capabilities, providers, policies, contexts, runtime resources, and evidence. The system evolves through discrete control and execution events, while physical time and resource usage remain continuous.

Notation in this section is scoped to the subsection in which it is defined unless explicitly carried forward. Object-model symbols from Section 6 (for example C, E, T, R, K, and H) are redefined here with the meanings stated at each use; where ambiguity could otherwise arise, distinct subscripts or multi-letter symbols such as Env_t and n_c are used.

## 12.2 System Sets and State

$$A=\{a_1,\ldots,a_n\},\ I=\{i_1,\ldots,i_m\} \quad (1)$$
$$C=\{c_1,\ldots,c_p\},\ P=\{p_1,\ldots,p_q\}$$

Here, $\mathcal{A}$ is the set of actors, $\mathcal{I}$ the set of intents, $\mathcal{C}$ the set of capability contracts, and $\mathcal{P}$ the set of provider implementations.

$$\Pi=\{\pi_1,\ldots,\pi_r\},\ X_t\in X \quad (2)$$
$$R_t\in R,\ E_t\in E$$

The remaining state variables identify policies, context, runtime resources, and accumulated evidence. The global architectural state at logical step t is then represented by Equation (3).

$$S_t=\left(I_t,C_t,P_t,\Pi_t,X_t,R_t,D_t,E_t\right) \quad (3)$$

In Equation (3), D_t is the delegation graph, and the time-indexed sets denote snapshots at logical step t of the corresponding sets defined in Equations (1) and (2). A federated deployment does not require any one component to hold the complete S_t; each domain maintains an authorized projection and exchanges only the state required by its federation contract.

## 12.3 Feasibility Before Optimization

For intent i and candidate provider p, each hard-constraint function h_j(i,p,S_t) returns 1 when the constraint is satisfied and 0 otherwise. A provider is feasible only when all of the n_c mandatory constraints evaluate to 1.

$$F(i,S_t)=\left\{p\in P \mid \prod_{j=1}^{n_c} h_j(i,p,S_t)=1\right\} \quad (4)$$

Hard constraints include authorization, policy, data residency, protocol compatibility, minimum trust, mandatory evidence, required hardware class, deadline feasibility, and side-effect compatibility. This separation is fundamental: utility optimization cannot override a failed hard constraint.

When F is empty, the control plane returns no feasible capability, requests clarification, relaxes only explicitly soft preferences, seeks human approval for an exception, or terminates. It must not silently convert a hard rule into a penalty.

## 12.4 Multi-Objective Provider Selection

For each feasible provider p, the model uses normalized attributes for semantic fit F_p, trust T_p, predicted confidence C_p, reliability R_p, locality G_p, latency L_p, cost K_p, energy E_p, and queue or congestion Q_p.

$$U(p\mid i)=w_f F_p+w_t T_p+w_c C_p \quad (5)$$
$$+w_r R_p+w_g G_p$$
$$-w_l L_p-w_k K_p-w_e E_p-w_q Q_p$$

The weights are intent- and policy-dependent. A latency-sensitive interactive request assigns a larger w_l. A regulated workflow assigns larger trust and risk weights. Scalarization is convenient but can hide tradeoffs; implementations may use lexicographic priorities, constrained optimization, Pareto frontiers, or learned policies, provided the decision remains explainable.

Utility terms must declare normalization and evidence. A provider-supplied “quality score” is not directly comparable to a calibrated enterprise evaluation. The selection record stores both raw evidence and transformed values.

## 12.5 Worked Selection Example

*Table 12. Illustrative capability-selection inputs. Values are hypothetical.*

| Candidate | Fit | Trust | Confidence | Reliability | Latency penalty | Cost penalty | Feasible? |
|---|---|---|---|---|---|---|---|
| **Provider A / us-west** | 0.92 | 0.95 | 0.87 | 0.99 | 0.20 | 0.40 | Yes |
| **Provider B / us-east** | 0.94 | 0.90 | 0.90 | 0.98 | 0.55 | 0.25 | No: residency |
| **Provider C / local** | 0.82 | 0.99 | 0.84 | 0.97 | 0.10 | 0.65 | Yes |

Provider B is removed before ranking because the intent requires us-west-only processing. Providers A and C remain. If quality and reliability dominate, A may be selected; if trust, local latency, or data-control preference dominates, C may be selected. The example illustrates why an apparently superior model or endpoint can be architecturally infeasible.

AOS records the rejected candidate and reason. This supports later analysis of whether policy, cost, or capacity—not model capability—caused the decision.

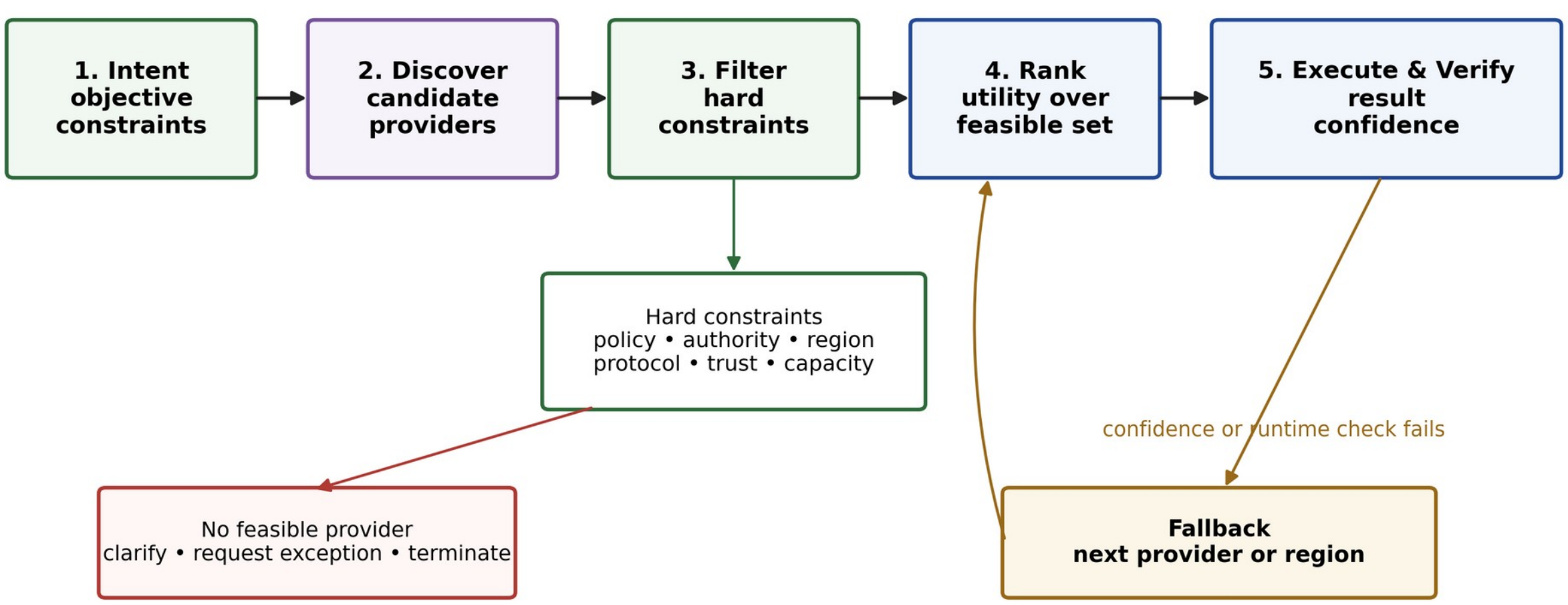

*Figure 8. Capability selection and fallback. Hard constraints define the feasible set before utility ranking; failed verification triggers bounded fallback rather than unrestricted retry.*

## 12.6 Delegation Attenuation and Authority Monotonicity

Let d_p denote a parent delegation and d_c a child delegation. Delegation is authority-monotone only when the child cannot expand the parent's permissions, scope, budget, or validity interval.

$$Perm(d_c) \subseteq Perm(d_p) \tag{6}$$

$$Scope(d_c) \subseteq Scope(d_p) \tag{7}$$

$$B(d_c) \leq B_{remaining}(d_p) \tag{8}$$

$$t_{end}(d_c) \leq t_{end}(d_p)$$

The subset relation is semantic: “read customer case 123” is narrower than “read all support records.” AOS profiles must define how scopes are compared. When a downstream actor needs broader authority, the system obtains a new grant from an independent authority rather than pretending the parent delegated it. When a parent creates multiple concurrent child delegations, the sum of the children’s budgets must not exceed the parent’s remaining budget, and budget accounting must be atomic.

Revocation creates a temporal problem. If an asynchronous task is disconnected, revocation may not be instantaneous. The delegation record therefore includes revocation status, lease or expiry, and the maximum tolerated revocation propagation delay. High-risk capabilities should use short-lived grants and online validation.

## 12.7 Confidence Aggregation

Let the confidence evidence contain k normalized components c_j in [0,1]. Each component has a non-negative weight α_j, and the weights sum to 1. Equation (9) defines one conservative aggregation using a weighted geometric mean; the floor ε is a small positive constant that prevents a zero-valued component from collapsing the aggregate to zero while still penalizing it strongly.

$$C_{agg}=\prod_{j=1}^{k} max(\varepsilon, c_j)^{\alpha_j} \quad (9)$$

$$\alpha_j \geq 0, \sum_{j=1}^{k} \alpha_j = 1$$

The geometric mean penalizes a very weak component more strongly than an arithmetic mean. It does not by itself solve dependence or calibration. If two components derive from the same evidence, treating them as independent double-counts information. AOS-0005 should define evidence groups, dependence annotations, calibration procedures, and uncertainty intervals. Whether a control decision gates on the point aggregate or on the lower uncertainty bound is likewise policy-defined; the canonical example in Appendix A gates on the point aggregate.

For a sequence of steps, a naive product of success probabilities can become unrealistically small or overconfident depending on dependence. The architecture therefore preserves step-level confidence and supports workflow-specific aggregation rather than mandating one universal formula.

## 12.8 Risk and Control Action

$$Risk(a)=Pr(H \mid e, x, a) \cdot Impact(H) \cdot Exposure(a) \quad (10)$$

In Equation (10), a is the candidate action; H denotes a harm event, local to this equation; e and x are the current evidence and context; Impact(H) is the severity of the harm; and Exposure(a) measures the scope of assets or subjects affected by the action. The formulation is schematic — a complete treatment sums expected harm over the relevant harm scenarios — and identifies the inputs a control policy must weigh. The control policy of Equation (11) maps aggregated confidence, risk, reversibility (Rev), the authority context (A), and applicable policy (Π) to an explicit runtime action u drawn from Table 13.

$$u=\pi_{control}(C_{agg}, Risk, Rev, A, \Pi) \quad (11)$$

*Table 13. Confidence- and risk-aware control actions.*

| Action | Typical condition | Required behavior |
|---|---|---|
| **PROCEED** | Confidence and policy thresholds satisfied | Execute within directive and standard evidence profile |
| **PROCEED_WITH_WARNING** | Residual uncertainty acceptable and action reversible | Execute with enhanced telemetry or operator notification |
| **RETRY** | Transient or recoverable uncertainty | Repeat with bounded change and attempt budget |
| **FALLBACK** | Primary provider or region insufficient | Choose next feasible provider and preserve lineage |
| **ESCALATE** | Human judgment or independent authority required | Pause and create accountable review task |
| **TERMINATE** | Unauthorized, unsafe, exhausted, or irrecoverable | Stop work, contain side effects, preserve evidence |

## 12.9 Scheduling and Head-of-Line Blocking

For an illustrative M/G/1 queue [43], λ is the request-arrival rate, S is the service-time random variable, and ρ = λE[S] is runtime utilization. The expected queueing delay is given by Equation (12).

$$E[W_q]=\frac{\lambda E[S^2]}{2(1-\rho)}, \rho=\lambda E[S] \quad (12)$$

The second moment $E[S^2]$ shows why long requests can increase waiting time for short requests even when average service time appears acceptable. Request-level FIFO can therefore produce head-of-line blocking. Token-level, step-level, shortest-remaining-processing-time, deadline-aware, or weighted-fair scheduling can reduce specific pathologies, though each introduces overhead and fairness tradeoffs.

AOS scheduling classes may include interactive, batch, safety-critical, human-blocked, deadline, and best-effort, drawing on established real-time scheduling practice [18]. The control plane provides priority and policy; the runtime scheduler selects an algorithm compatible with the provider and infrastructure.

## 12.10 Latency and Placement

$$\begin{aligned} D_{total} = D_{control} + D_{discovery} + D_{queue} \\ + D_{network} + D_{inference} + D_{tool} \\ + D_{validation} + D_{audit} \end{aligned} \tag{13}$$

This decomposition prevents the system from treating model inference latency as the entire user-visible delay. Tail latency matters. A provider with a lower mean can be inferior if queueing or network variance creates an unacceptable P99 [17].

Physics imposes a non-negotiable propagation bound. For path length d and signal velocity v in fiber, propagation delay is at least d/v, with v approximately $2 \times 10^8$ meters per second before routing, serialization, switching, and queueing delay are added.

A placement problem can be expressed by minimizing an explicit objective J(p) over the feasible provider set:

$$\begin{aligned} \underset{p \in F(i, S_t)}{minimize}\, J(p) \\ J(p) = w_L D_p + w_K K_p + w_E E_p \\ + w_Q Q_p - w_R R_p \end{aligned} \tag{14}$$

subject to policy, authority, residency, capacity, protocol, and deadline constraints. In Equation (14), D_p denotes the predicted end-to-end delay of provider p per the decomposition of Equation (13), and K_p, E_p, Q_p, and R_p are the cost, energy, congestion, and reliability attributes of Section 12.4. The objective can be solved centrally, hierarchically, or locally depending on the deployment profile.

## 12.11 Cost, Energy, and Carbon

$$\begin{aligned} C_{total} = C_{model} + C_{tool} + C_{data} \\ + C_{network} + C_{human} + C_{control} \end{aligned} \tag{15}$$

$$\begin{aligned} E_{energy} = \int_0^T P(t)\, dt \\ C_{carbon} = \int_0^T P(t)\, g(t)\, dt \end{aligned} \tag{16}$$

In Equation (16), P(t) is the instantaneous power draw of the execution, g(t) is the grid carbon intensity at time t, and T is the execution interval. Economic reliability means that cost remains predictable enough for sustained operation. Token price is only one component; retries, tool calls, data movement, observability retention, human review, and idle reserved capacity can dominate. AOS budgets can be allocated per intent and delegated monotonically to children.

Energy-aware routing may defer flexible work, select a more efficient provider, or place computation near data [46]. These decisions must remain subordinate to safety, authority, and correctness. A cheaper or lower-carbon path that violates data residency is infeasible.

## 12.12 Evidence-Driven Control

Let S_t* denote the desired semantic state and S_t the observed state. The control policy chooses action u_t using the state residual, evidence, and applicable policy.

$$u_t = \pi \left( S_t^{\star}, S_t, E_t, \Pi_t \right) \tag{17}$$

The runtime then applies the control action, advances the system through its dynamics, and produces a new state and new evidence.

$$\left( S_{t+1}, E_{t+1} \right) = F \left( S_t, u_t, \omega_t \right) \tag{18}$$

In Equation (18), ω_t represents exogenous disturbances such as provider failures, input variation, and environmental change. The formulation is intentionally general because semantic states are partly discrete and partly continuous. The control-loop analogy is nevertheless useful [42]: AOS measures deviation between intended, authorized, observed, and resulting state; applies bounded actions; and updates evidence. Stability in this context means avoiding unbounded oscillation such as retry storms, repeated replanning, delegation loops, or policy thrashing.

Control policies therefore include hysteresis, attempt budgets, circuit breakers, backoff, and escalation. A future formal model could use hybrid systems or temporal logic to verify selected properties.

The evidence-driven reconciliation cycle is shown in Figure 3. Equations (17) and (18) formalize the control action and resulting state transition.

## 12.13 Time, Ordering, and Causality

Distributed agentic execution requires both physical time and logical causality [60]. Wall-clock timestamps support latency, deadlines, token validity, and audit chronology, but they do not by themselves prove causation.

A timestamp may be represented as the uncertainty interval $[t - \varepsilon, t + \varepsilon]$. If the upper bound of event A precedes the lower bound of event B, physical time supports the ordering A before B. If the intervals overlap, ordering must come from logical clocks, message dependencies, trace parentage, or an explicit causal edge.

This distinction matters for forensic analysis. A policy event may appear later than an execution event because of clock skew even though the execution depended on the policy. Causal edges preserve the logical relationship.

## 12.14 Lifecycle State Machine

The canonical lifecycle exposes externally meaningful states while allowing runtime-specific substates. Every transition emits an event and preserves monotonic audit history. Retry and fallback return to bounded earlier states without erasing the failed attempt.

Terminal states distinguish completed, terminated, and failed. Completed means the declared success criteria were met. Terminated means policy or authority deliberately stopped the work. Failed means the runtime could not complete and did not select a permitted recovery. Escalated may remain non-terminal while waiting for human action.

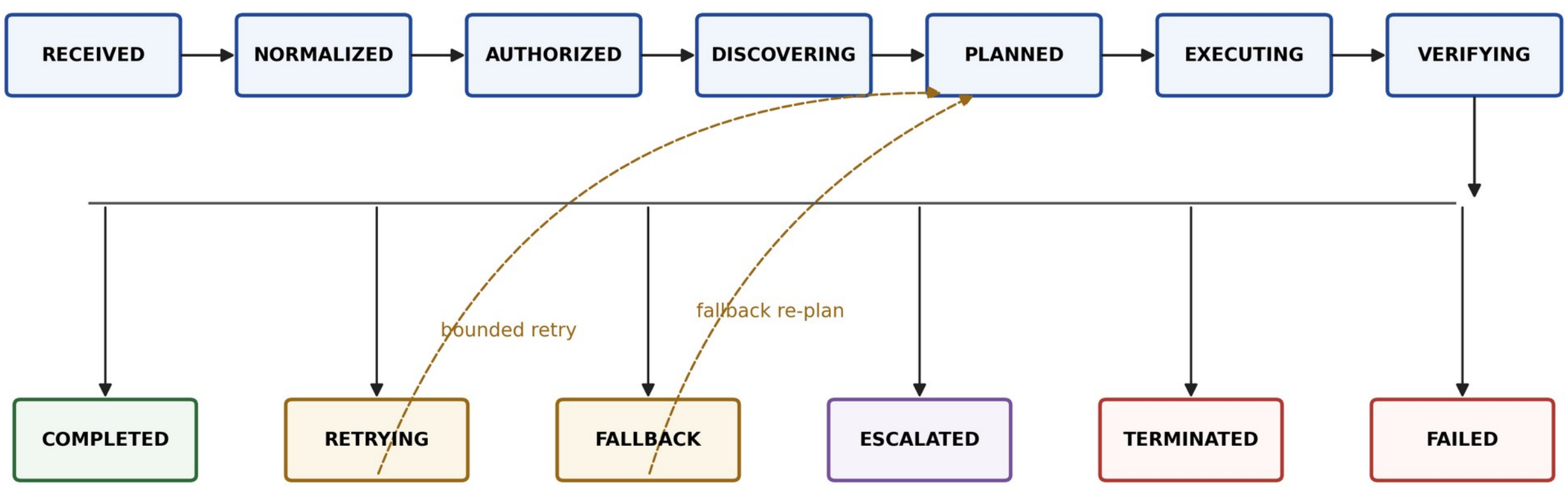

*Figure 9. Canonical AOS intent and execution lifecycle with explicit terminal outcomes and bounded retry or fallback transitions.*

# 13. Deployment Models

## 13.1 Deployment Principles

AOS logical planes are independent of physical topology. The same architecture can run as an embedded library, a central service, a regional control domain, a federated global system, an edge hierarchy, or a sovereign isolated deployment. The profile determines availability, consistency, trust, latency, and administrative tradeoffs.

A deployment declaration identifies where authoritative intent, policy, registry, delegation, runtime state, and audit evidence reside; how failures are handled; and which cross-domain claims are trusted. Without these declarations, diagrams showing “global AI orchestration” conceal important control assumptions.

## 13.2 Embedded Profile

In the embedded profile, control and runtime modules execute inside one application process or host. The registry may be a local file or in-memory store, and adapters call local functions or remote services. This profile minimizes latency and operational complexity and is suitable for development, constrained devices, and single-application deployments.

The tradeoff is limited fault isolation and shared fate. An application defect can affect control and execution simultaneously. Persistent audit and external policy services may still be required for high-risk use cases. Embedded conformance should not be confused with full enterprise availability.

## 13.3 Centralized Control Profile

A centralized profile uses one logical control service for a cluster or enterprise domain and one or more runtime workers. It provides consistent policy, a unified registry, and simpler audit. The control service can scale horizontally and use a replicated state store, much like established cluster control planes.

The main risks are control-plane availability, blast radius, and policy bottlenecks. The architecture requires cached safe behavior, idempotent directives, leader or quorum failure handling, and clear behavior when the control plane is unreachable. Runtime components must not interpret unavailability as permission to bypass control.

## 13.4 Regional Profile

A regional profile places control and runtime planes near users, data, and compute. Each region can enforce local residency, latency, and availability policy while receiving global policy templates and capability metadata. Regional state reduces WAN dependence for routine decisions.

The challenge is consistency. Capability health, policy revocation, identity, and delegation state can lag across regions. Profiles define which data is strongly consistent, which is eventually consistent, and which decisions require an online global authority.

## 13.5 Federated Multi-Region and Multi-Organization Profile

Federation allows independent AOS domains to retain local authority while exchanging signed capability advertisements, policy claims, trust relationships, and execution evidence. A global registry need not own every record; it may index domain registries and expose verifiable references. Cross-region routing occurs only when both source and destination policies permit it.

Federation introduces policy conflicts, identity mapping, revocation latency, evidence disclosure, and economic settlement. The architecture therefore separates capability discovery from authority. Discovering a remote capability does not grant permission to invoke it. Negotiation may establish a bounded cross-domain delegation and evidence contract.

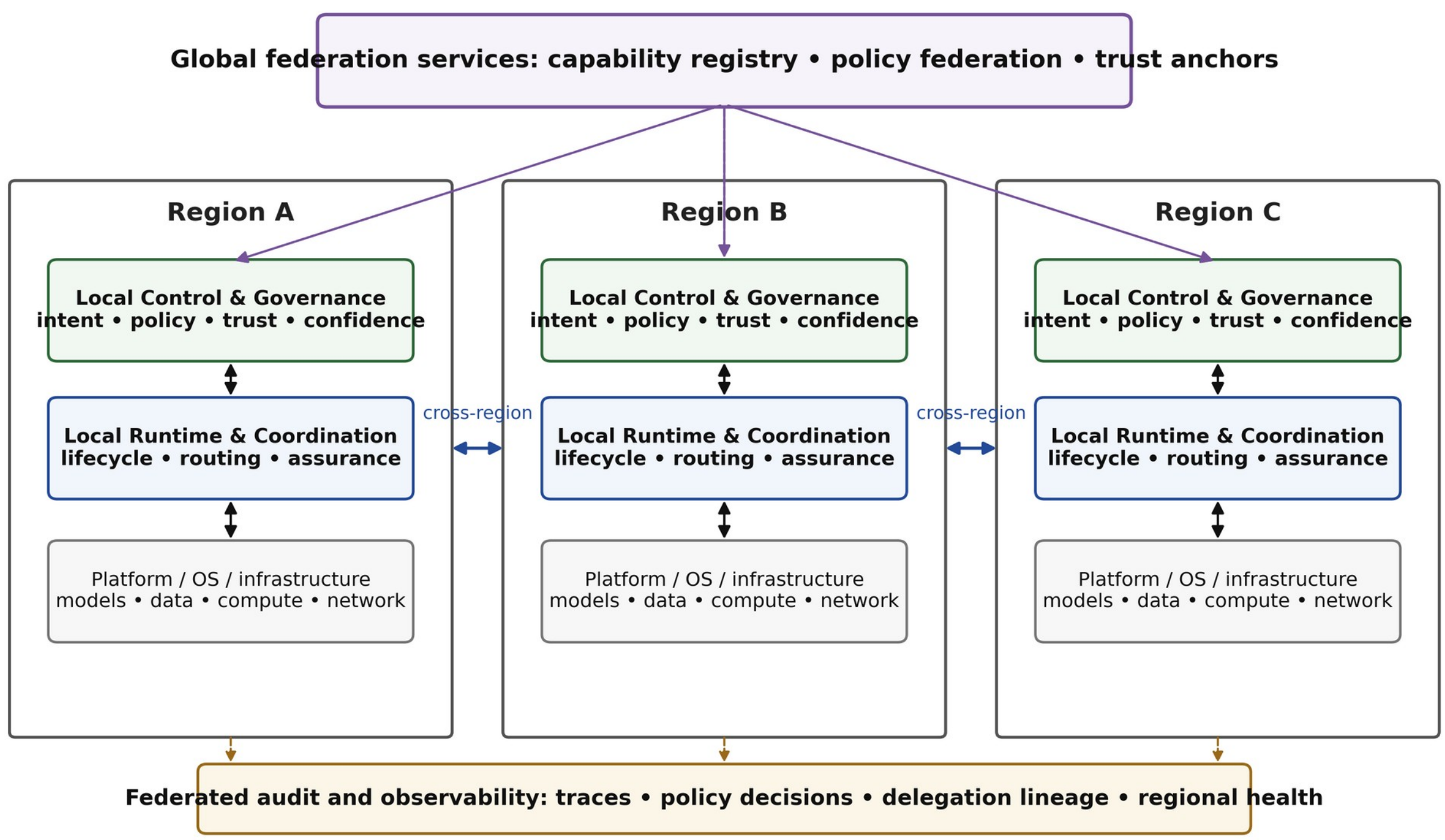

*Figure 10. Federated multi-region AOS. Regional control and runtime domains share capability and policy federation while retaining local execution and failure containment.*

## 13.6 Edge and Distributed Profile

Edge deployments place runtime and selected control functions near sensors, users, devices, or local data. A central or regional plane distributes policy, capability metadata, and models; the edge performs low-latency coordination and can continue during intermittent connectivity.

Edge constraints include limited compute, power, storage, bandwidth, and physical security. AOS profiles may use reduced models, local deterministic rules, short-lived delegations, compact evidence, and delayed audit upload. Offline operation must be bounded by lease, capability, and risk class.

## 13.7 Sovereign and Air-Gapped Profile

A sovereign profile keeps intents, models, data, registry state, policy, and evidence inside a defined administrative or regulatory boundary. External capabilities are prohibited or imported through controlled review. Trust roots and identity are local, and observability data does not leave the domain.

The profile improves autonomy and compliance but reduces provider diversity and rapid external update. Model, policy, and vulnerability updates require a governed import process. Cross-domain collaboration uses explicit gateways or sanitized artifacts rather than implicit cloud calls.

## 13.8 Hybrid Profile

Most enterprises will use hybrid deployments. A central policy authority may coexist with regional registries, local runtime schedulers, sovereign data zones, and cloud model providers. AOS makes the hybrid structure explicit by declaring authority, registry, runtime, and evidence boundaries.

Hybrid design should minimize synchronous dependencies across wide-area links. Local domains need enough policy and capability state to make safe decisions during transient partitions, while high-risk or cross-border actions may require online coordination.

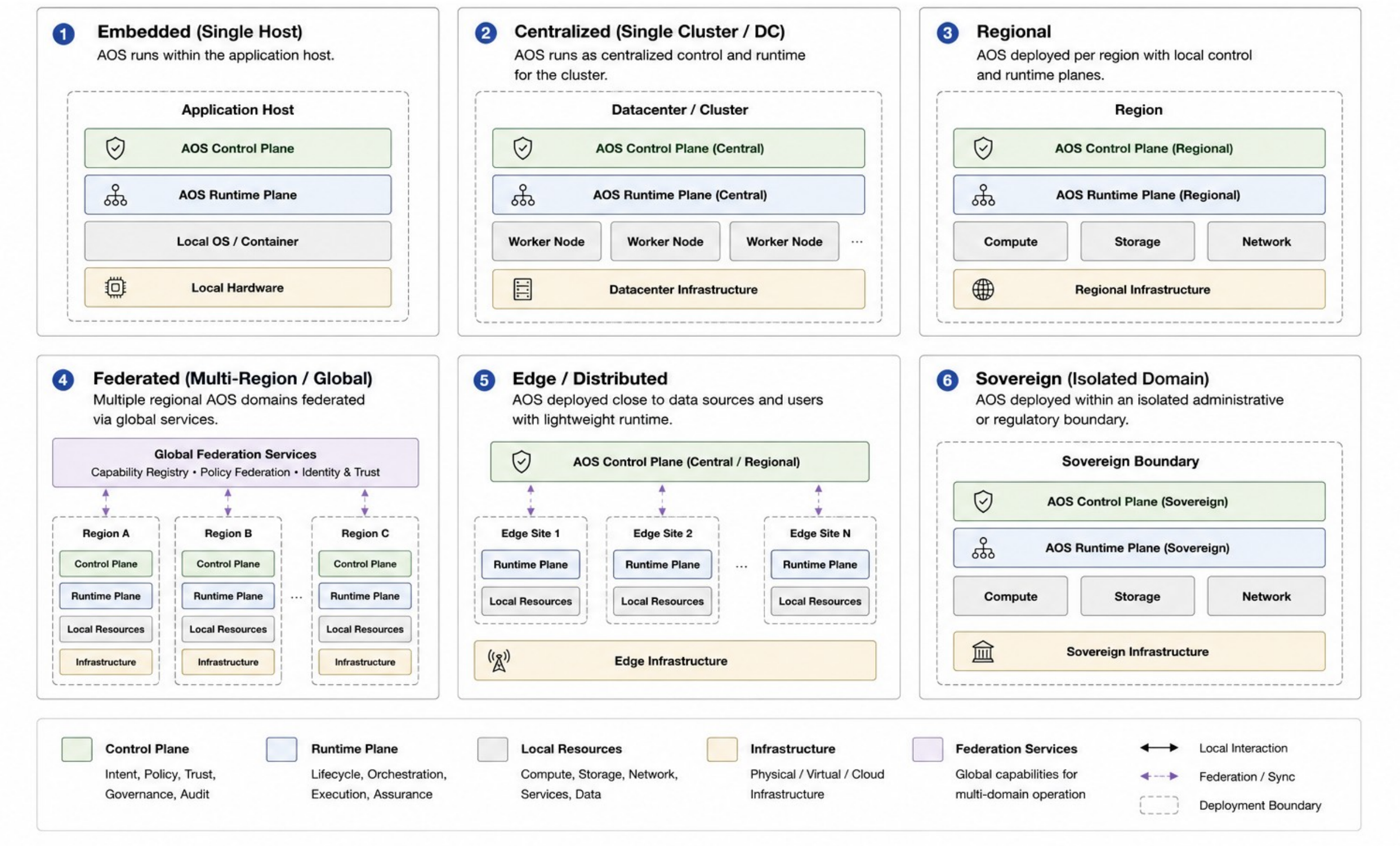

*Figure 11. Deployment profiles: embedded, centralized, regional, federated, edge, and sovereign.*

## 13.9 Linux and Windows Portability

The AOS semantic model is independent of host operating system. Linux and Windows adapters translate runtime constraints into native process, identity, filesystem, networking, container, and telemetry mechanisms. Cross-OS federation occurs at AOS and platform interfaces rather than by pretending the kernels are identical.

Portability requires declared capability, not lowest-common-denominator behavior. A provider may advertise that it supports a particular sandbox, workload identity, accelerator, or audit mechanism in only one environment. The control plane uses this metadata during feasibility evaluation.

# 14. Reliability

## 14.1 Reliability as a System Property

AI reliability is the ability of a system to achieve intended outcomes consistently while remaining within defined performance, safety, security, trust, governance, and economic constraints. It is not equivalent to model accuracy. A highly accurate model can participate in an unreliable system if context is stale, tools are unavailable, policy is inconsistent, delegation is opaque, or infrastructure is unstable.

AOS treats reliability as an end-to-end property produced across the control plane, runtime plane, platform services, operating system, infrastructure, and human operation. Each layer has distinct mechanisms and metrics. The control plane governs requirements and response; lower layers produce execution and evidence.

**Reliability Is Produced Across Layers and Governed End to End**

Control & Governance
policy, trust, confidence, audit

Runtime & Coordination
lifecycle, scheduling, routing, assurance

Platform Services
models, data, identity, messaging

OS / Kernel
processes, files, sockets, isolation

Infrastructure
compute, memory, network, storage, power

End-to-End Reliability

*Figure 12. Reliability is produced across control, runtime, platform, operating-system, and infrastructure layers and governed end to end.*

## 14.2 Reliability Dimensions

*Table 14. End-to-end AI reliability dimensions.*

| Dimension | Meaning | Example evidence |
|---|---|---|
| **Functional** | Correctness, task success, factuality, completeness, constraint satisfaction | Success criteria, verifiers, outcome checks |
| **Availability** | Service readiness and recoverability | Uptime, MTBF, MTTR, failover success |
| **Performance** | Predictable latency, throughput, and resource behavior | P50/P95/P99, queue delay, tokens/s, task/s |
| **Data and context** | Fresh, complete, authorized, and traceable information | Freshness age, retrieval precision/recall, lineage completeness |
| **Robustness** | Graceful behavior under OOD input, noise, overload, and adversarial conditions | Degradation curves, error containment, fallback success |
| **Model stability** | Controlled drift and regression across versions | Evaluation deltas, calibration drift, rollback time |
| **Safety** | Avoidance of harmful or prohibited outcomes | Violation rate, prevented side effects, escalation effectiveness |

| Dimension | Meaning | Example evidence |
|---|---|---|
| **Security** | Confidentiality, integrity, availability, and containment | Prompt-injection resistance, credential misuse, isolation breach rate |
| **Explainability and observability** | Ability to reconstruct state, decisions, and causal paths | Trace completeness, decision coverage, forensic query success |
| **Fairness and bias** | Consistent treatment and performance across relevant populations | Subgroup error, parity measures, policy-defined thresholds |
| **Agentic** | Reliable multi-step planning, delegation, coordination, and termination | Loop rate, task completion, delegation violations, recovery success |
| **Economic** | Predictable cost and resource efficiency | Cost/task, variance, budget violations, energy/task |

## 14.3 End-to-End Success and Dependency Structure

If a workflow contains n serial components with independent success probabilities r_i, a simplified series reliability is:

$$R_{series} = \prod_{i=1}^{n} r_i \tag{19}$$

The independence assumption is often false because components share models, data, infrastructure, or policy. The formula nevertheless illustrates why adding steps without verification and recovery can reduce end-to-end success. AOS preserves dependency and common-cause information so reliability analysis can avoid naive multiplication.

For two independent redundant providers with success probabilities r_1 and r_2, the probability that at least one succeeds is 1 - (1-r_1)(1-r_2). Redundancy only improves reliability if failures are sufficiently independent. Two providers in the same region, using the same model and data source, may share a common cause.

## 14.4 Availability and Recovery

$$A = \frac{MTBF}{MTBF + MTTR} \tag{20}$$

Availability metrics apply to registries, policy evaluators, runtimes, model endpoints, data systems, telemetry, and human approval services [56]. AOS additionally measures recovery correctness: a fast failover that violates policy or duplicates a side effect is not reliable.

Recovery actions include retry, resume from checkpoint, provider fallback, regional failover, compensation, degraded mode, human escalation, and termination. The failure contract identifies which actions are safe. Recovery is bounded by attempt, time, cost, and authority budgets.

## 14.5 Functional Reliability and Verification

Functional reliability is evaluated against explicit success criteria rather than generic model quality. A summarization capability may require source coverage, citation validity, schema correctness, and absence of prohibited data. A configuration capability may require a dry run, policy check, invariant validation, and post-change health confirmation.

Verification can use deterministic tests, independent models, retrieval checks, domain rules, simulation, human review, or downstream outcomes. AOS records which verifier was used and its own trust and confidence. Verification is part of the operating path, not an optional evaluation performed after deployment.

## 14.6 Context and Memory Reliability

Many apparent reasoning failures are context failures: the system retrieved the wrong record, omitted a constraint, used stale memory, or exposed irrelevant history. Context reliability therefore includes source selection, provenance, freshness, authorization, contradiction handling, supersession, and scope.

AOS distinguishes contextual state from authoritative state. A memory record can suggest a user preference, but policy or a current explicit instruction may override it. Contradictory records are not averaged silently; the system preserves version history and applies a resolution rule or requests clarification.

## 14.7 Robustness and Graceful Degradation

Robust systems degrade predictably under overload, provider loss, adversarial input, distribution shift, and partial data. AOS defines degraded profiles, such as local deterministic rules when remote models are unavailable, reduced capability sets during an incident, or human review when confidence drops.

Graceful degradation is policy-specific. A customer-support assistant may provide a limited answer; a safety-critical controller may halt. The runtime records the degraded mode and prevents it from being mistaken for normal operation.

## 14.8 Safety, Security, and Kernel Enforcement

Safety and security overlap but are not identical. Safety focuses on harmful outcomes; security on unauthorized access, manipulation, disclosure, and disruption. AOS policy can prohibit capabilities, require approval, limit data, and select trusted providers. Runtime assurance can sandbox processes, restrict networks, rate-limit tools, and terminate behavior.

The final local enforcement boundary may reside in the operating system or hypervisor. High-level authority can be translated into filesystem, process, namespace, network, device, and syscall constraints. The translation must be conservative: the kernel need not understand the human objective, but it can enforce a bounded execution envelope derived from the approved authority chain.

## 14.9 Observability and Delegation Reliability

Reliability analysis requires semantic observability. Operators must know which intent and policy were active, which capability and provider were selected, what context was used, how authority was delegated, which confidence changed, and what side effects occurred. Conventional metrics alone cannot answer these questions.

A useful quality metric is semantic trace completeness: the fraction of consequential actions that can be linked to a valid intent, authority, policy decision, capability, provider, delegation, and result. Another is delegation reconstruction success: whether an investigator can enumerate all actions and data access under a delegation without heuristic time-window correlation.

## 14.10 Economic Reliability

An agentic system that succeeds technically but produces unpredictable cost is operationally unreliable. Economic reliability includes budget adherence, cost variance, retry amplification, unused reserved capacity, observability retention, human review, and energy. The control plane sets budgets and objectives; the runtime meters actual use and applies backpressure or fallback.

Cost policy should account for failure cost. Selecting a cheaper but less reliable provider can increase total cost through retries, human review, or downstream error. AOS therefore treats economic optimization as part of multi-objective selection rather than a simple lowest-price rule.

## 14.11 Reliability Control Loop

1. Define intent-specific success, safety, security, performance, and cost requirements.
2. Select capabilities and providers whose evidence satisfies hard constraints and reliability profile.
3. Observe execution using semantic events, platform telemetry, and verification.
4. Assess confidence, risk, drift, and divergence from authorized state.
5. Apply bounded actions: proceed, warn, retry, fallback, isolate, escalate, compensate, or terminate (the Table 13 control actions extended with containment and compensation).
6. Record evidence and update provider health, trust, and reliability estimates.

The loop resembles NIST Govern, Map, Measure, and Manage at an architectural runtime level [5]. Organizational governance remains necessary; AOS supplies the objects and interfaces through which selected controls and evidence can be operationalized.

The planned AOS-0006 document is intended to define the reliability framework in greater detail, including maturity levels, metrics, evaluation datasets, failure injection, and reliability-focused conformance tests.

## 14.12 Measurement and Evaluation Plan

*Table 15. Proposed evaluation methodology.*

| Evaluation area | Method | Representative measure |
|---|---|---|
| **Architecture conformance** | Static declarations and contract tests | Required objects, interfaces, invariants, unsupported features |

| Evaluation area | Method | Representative measure |
|---|---|---|
| **Functional workflows** | Scenario suites with deterministic success criteria | Task success, constraint satisfaction, citation validity |
| **Failure recovery** | Fault injection and chaos experiments | Recovery success, duplicate side effects, MTTR, evidence completeness |
| **Scheduling** | Mixed short/long workload experiments | P50/P99 latency, fairness, throughput, starvation |
| **Federation** | Partition, stale state, and revocation experiments | Policy conflict rate, revocation delay, safe local operation |
| **Delegation observability** | Forensic reconstruction queries | Delegation-scoped attribution and data-footprint completeness |
| **Confidence control** | Calibration and decision-outcome studies | ECE/Brier score, unnecessary escalation, unsafe proceed rate |
| **Economic reliability** | Budget and workload variability tests | Cost/task, variance, budget violation, retry amplification |

# 15. Discussion

## 15.1 Why a Reference Architecture Rather Than Another Platform

The agentic AI ecosystem does not lack software. It contains many frameworks, model gateways, workflow engines, memory products, security layers, data platforms, and infrastructure services. The missing artifact is a stable way to describe how these components divide responsibility and interoperate. A product inevitably optimizes for its own deployment model; a reference architecture can classify several products without requiring any one of them to become universal.

This distinction changes the adoption model. An enterprise can implement AOS responsibilities using existing technology, purchase components from several vendors, or adopt aos-core selectively. A research project can contribute a scheduler or confidence method without implementing policy and identity. A protocol can bind to an AOS interface without redefining the architecture. The value lies in interoperability, comparability, and explicit gaps.

The architecture also reduces category confusion. A model and data platform, an agent kernel, a workflow framework, and a governance product may all market themselves as an “AI OS,” yet implement different responsibilities. AOS-0001 provides a matrix for describing the implemented plane, external dependencies, and unsupported functions. The term becomes an architectural claim rather than a branding claim.

## 15.2 Why Two Internal Planes

An earlier AOS interpretation placed only the control plane inside AOS and treated the agent runtime as a separate AIOS layer. The two-plane model is stronger for three reasons. First, end-to-end operating responsibility includes runtime lifecycle, scheduling, routing, context, and assurance. Second, many failures arise from the translation between approved intent and actual execution; excluding runtime coordination would leave this translation undefined. Third, AOS can still remain integration-oriented because Plane 2 coordinates external engines rather than replacing them.

The two-plane model does create scope risk. If every model server, workflow engine, memory database, and infrastructure scheduler is labeled part of AOS, the boundary becomes meaningless. The governing rule is therefore responsibility-based: AOS owns lifecycle and coordination semantics but may delegate execution mechanisms to providers. Platform services remain external even when tightly managed.

## 15.3 Control-Plane Centralization Versus Federation

Central control simplifies policy, registry state, and audit but creates availability, latency, concentration, and sovereignty concerns. Federation preserves local authority and supports multi-organization operation, but introduces stale metadata, trust negotiation, conflict resolution, and revocation delay. Neither model is universally superior.

AOS supports hierarchical composition. A local domain can make low-risk decisions from cached policy and capability state while consulting a regional or global authority for cross-domain, high-impact, or ambiguous actions. The profile defines which decisions require synchronous coordination. This resembles established distributed-systems designs in which local autonomy is bounded by leases, epochs, or quorum rules.

Research is needed to determine the right consistency model for each object. A policy revocation may require stronger consistency than provider latency telemetry. An authority grant may use short leases; a capability description may be eventually consistent. Treating all state identically would either create unnecessary latency or unsafe staleness.

## 15.4 Semantic Abstractions and Ontology Risk

Capability-centric design requires shared semantics. If two providers both advertise ai.case.summary.generate but interpret “case” or “cited” differently, discovery creates false interoperability. AOS therefore needs contract schemas, semantic versioning, side-effect declarations, evaluation profiles, and domain ontologies.

Over-standardization is also risky. A universal ontology for all human objectives is unrealistic and could slow innovation. The architecture separates globally stable concepts from domain-specific profiles. Intent, capability, authority, delegation, confidence, and evidence remain foundational; healthcare, telecom, cybersecurity, finance, and data-center profiles define specialized terms and constraints.

## 15.5 Confidence and the Danger of False Precision

Confidence is useful precisely because agentic systems are uncertain and model outputs remain subject to hallucination and miscalibration [31], but numerical values can create misleading precision. A score of 0.87 may reflect model calibration, a heuristic, or a subjective provider claim. AOS therefore requires evidence, evaluator identity, calibration scope, and uncertainty representation. The action mapping is more important than the number.

Confidence also interacts with dependence. Independent verifiers can improve assurance; two evaluations using the same model and context may not. A future confidence framework should represent source dependence and distinguish epistemic uncertainty, aleatoric uncertainty, policy certainty, and runtime health. Until then, implementations should avoid aggregating incomparable scores without documentation.

The architecture deliberately allows qualitative and interval-based confidence. A provider may report high/medium/low with evidence, or a calibrated interval, provided the control policy can interpret it. Conformance should focus on reproducibility and action semantics rather than one score format.

## 15.6 Human Oversight Is an Operating Function

Human-in-the-loop is often treated as a fallback UI. In practice, human authority is part of the operating system. Humans set policy, approve exceptions, interpret ambiguous objectives, resolve conflicting evidence, and accept responsibility for high-impact actions. Their availability, expertise, workload, and response time are schedulable resources.

AOS therefore models human roles and review tasks explicitly, consistent with established guidance for human-AI interaction [47]. This has organizational consequences: an enterprise must define who can approve what, what evidence is shown, how appeals work, and how emergency stops are exercised. Technical architecture cannot manufacture accountability that the organization has not assigned.

## 15.7 Security and Authority Translation

The most difficult security problem is translation from semantic authority to concrete enforcement. A policy may permit an agent to “read project logs” but the operating system requires paths, file descriptors, credentials, processes, sockets, and syscalls. Over-broad translation defeats least authority; under-broad translation prevents useful work.

AOS-0001 defines the authority chain and enforcement interface but does not solve general semantic compilation. Domain-specific capability manifests can help by declaring expected resources and side effects. Runtime sandboxes can learn a profile in controlled environments, but learned policies require review to avoid normalizing malicious or accidental behavior. Kernel assurance remains a complementary lower-layer mechanism rather than a replacement for high-level governance.

## 15.8 Observability, Privacy, and Evidence Cost

Semantic observability improves accountability but increases data volume and privacy risk. Context references, model inputs, delegation records, and provider decisions may contain sensitive information. AOS profiles should separate identifiers from protected payloads, apply minimization, encrypt evidence, define retention, and restrict forensic access.

Complete recording is not always feasible. The control plane can select evidence profiles based on risk. Low-risk actions may use sampled telemetry; high-impact actions may require full lineage and output hashes. Sampling decisions themselves become policy and audit events. The system should be explicit about which questions can and cannot be answered from retained evidence.

## 15.9 Scalability and Control-Plane Economics

Control adds latency and cost. Registry lookup, policy evaluation, trust verification, confidence assessment, and audit persistence consume resources. A poorly designed control plane can become the bottleneck it was intended to manage. Caching, local decisions, batched evidence, incremental evaluation, and risk-based control are therefore necessary.

The architecture supports fast paths. A low-risk, frequently repeated capability can use a cached authorization lease and local provider ranking, while high-risk or novel work uses the complete path. Fast paths remain bounded by policy version, expiry, context class, and revocation. Performance optimization cannot convert control into an optional best effort.

## 15.10 Limitations of This Work

- The paper proposes an architecture and mathematical model but does not present a large-scale empirical evaluation.
- Capability semantic interoperability is specified at a high level; domain ontologies and negotiation protocols remain future work.
- The invariants are conditional on validated capability metadata: whether an action is consequential depends on declared side effects, so detection of undeclared side effects falls to runtime assurance and lower-level enforcement, and high-assurance profiles should not rely on declaration alone.
- Confidence aggregation is intentionally under-specified because calibration and dependence vary by domain.
- Federated policy conflict resolution and revocation consistency are open research problems.
- The architecture assumes that external platforms expose sufficient control and telemetry; legacy systems may require invasive adapters.
- Human governance, legal accountability, and organizational incentives cannot be solved by technical architecture alone.
- Some responsibilities may move between planes as implementation experience accumulates; the lead-editor and ADR process exists to manage this evolution.

### 15.11 Validation Strategy

The architecture should be validated through three complementary methods. First, aos-core and independent implementations can demonstrate that canonical objects, SPIs, delegation, confidence actions, and audit semantics are implementable. Second, scenario-based tests can evaluate normal operation, policy denial, provider failure, regional fallback, human escalation, and revocation. Third, landscape mapping can test whether the architecture classifies existing protocols and products without forced or ambiguous placement.

Empirical validation should compare not only throughput but also decision traceability, policy enforcement, recovery correctness, delegation reconstruction, cost predictability, and operator comprehensibility. A system that is faster but cannot explain or bound authority may not be superior for enterprise use.

# 16. Future Work

## 16.1 Intent Algebra and Formal Semantics

A formal intent language could define composition, refinement, conflict, partial satisfaction, and equivalence. Such an algebra would help determine whether a plan remains faithful to the original objective and whether two intents can share work or resources. It must preserve human interpretability and avoid becoming a domain-specific programming language disguised as natural language.

Temporal logic may express deadlines, ordering, persistence, and forbidden intervals. Deontic logic may help represent obligation, permission, and prohibition. Practical profiles will likely combine formal constraints with natural-language objective descriptions.

## 16.2 Capability Semantics, Registries, and Negotiation

AOS-0004 should define capability naming, schema, side-effect taxonomy, health, trust evidence, evaluation profiles, federation, and negotiation. Open questions include how to verify provider claims, how to represent composite capabilities, and how to avoid registry spam or strategic metadata.

Negotiation may include price, latency, data-use terms, evidence obligations, and delegated authority. This creates a potential market for capability liquidity, but economic optimization must remain subordinate to policy and trust.

## 16.3 Confidence Calculus and Runtime Assurance

AOS-0005 should develop calibrated confidence records, dependence-aware aggregation, uncertainty intervals, decay, propagation, and action policies. Research is needed to combine model calibration, retrieval quality, provider reliability, rule-based verification, and human judgments without false precision.

Runtime assurance can use monitors, shadow execution, independent verifiers, conformal methods [48], control barriers, or formal invariants. The architecture should support pluggable assurance while making the cost and evidence visible.

## 16.4 Scheduling and Traffic Management

Agentic scheduling requires algorithms for variable-length generation, tool wait, human wait, multi-agent fan-out, shared memory, and heterogeneous accelerators. Research should compare request-, token-, step-, deadline-, confidence-, and risk-aware scheduling under realistic workload distributions.

Traffic management must control retry amplification, cross-region routing, provider quotas, circuit breakers, and partial side effects. It may benefit from network and topology awareness, especially for data-intensive and collective workloads.

## 16.5 Federation, Time, and Global Coordination

Federated AOS domains require trust negotiation, policy translation, registry consistency, economic settlement, and evidence exchange. Short-lived leases and signed claims can bound stale authority, but wide-area failure and disconnected operation remain difficult.

Time synchronization and uncertainty should become explicit scheduling and audit inputs. Research should quantify when physical-time guarantees are required, when logical causality is sufficient, and how timing failures affect confidence and forensic reconstruction.

## 16.6 Memory, Context, and State Governance

Long-term agent memory raises questions of ownership, correction, contradiction, supersession, retention, and deletion. AOS should define memory-provider contracts and evidence requirements without standardizing one storage technology. Context selection should be testable because reliability often depends on what the model did or did not see.

Causality-aware memory could record which facts influenced which decisions. This would improve audit and permit targeted invalidation when a source is corrected.

## 16.7 Proof of Human, Proof of AI, and Actor Attestation

As humans and agents become difficult to distinguish at the interface, services may require proof of human, proof of agent, proof of device, proof of organization, or proof of execution environment. AOS can consume these attestations as trust inputs while keeping authorization separate.

Research is needed to balance accountability with privacy. A pseudonymous human may be sufficiently authorized for one capability, while a regulated transaction may require strong identity and non-repudiation.

## 16.8 Formal Verification and Model Checking

Selected AOS invariants can be expressed in TLA+, temporal logic, or state-machine models [57]. Candidate properties include non-expanding delegation, no execution without authorization, eventual termination under bounded retry, audit monotonicity, revocation safety, and exactly-once side-effect profiles.

Formal methods will not verify open-ended model reasoning, but they can verify the deterministic control and lifecycle boundaries around it. This is a practical division of labor between probabilistic intelligence and formal assurance.

## 16.9 Conformance Suite, Landscape, and Maturity Model

AOS-0010 should provide executable examples and conformance tests for canonical objects, adapters, policy decisions, delegation, confidence actions, and semantic events, building on the conformance vocabulary to be defined in AOS-0002 and the reliability evaluation methods planned for AOS-0006. Tests should be profile-based rather than requiring every implementation to supply every function.

A public landscape can allow projects, protocols, products, and research systems to register the capabilities and layers they implement. A maturity model can help enterprises assess architecture, governance, reliability, security, observability, and operational readiness. These community artifacts should remain distinct from the aos-core implementation.

## 16.10 Empirical Research Program

A research program for AOS should include:

- mixed-workload scheduling benchmarks with short and long generation, tools, and human wait;
- multi-region experiments measuring tail latency, policy constraints, revocation, and fallback;
- delegation-observability studies measuring forensic reconstruction and data-access attribution;
- confidence calibration and action-policy experiments across reversible and high-impact tasks;
- fault-injection studies covering registry, policy, provider, network, telemetry, and identity failures;
- operator studies measuring comprehensibility, incident response, and trust in the control plane;
- economic experiments comparing lowest-cost routing with reliability- and risk-aware routing.

## 17. Conclusion

Agentic AI systems are becoming distributed operating environments. They combine probabilistic reasoning, dynamic tool use, memory, delegation, human authority, and external side effects across heterogeneous runtimes and infrastructure. The ecosystem has produced many valuable execution technologies, but it lacks a shared architecture that relates intent, authority, capability, policy, confidence, runtime coordination, and evidence.

This paper proposed the Agent Operating System as a vendor-neutral reference operating architecture with two internal planes. The Control & Governance Plane determines what should happen, under whose authority, and subject to which constraints. The Runtime & Coordination Plane determines how approved work is coordinated and assured. Platform services, Linux or Windows, and physical infrastructure remain external but managed through explicit interfaces.

The foundational contribution is not a claim that one implementation should own the entire stack. It is a set of stable responsibilities and invariants that allow several implementations to interoperate and be compared. Capability-centric discovery, authority-preserving delegation, actionable confidence, semantic observability, and end-to-end reliability form the core of that model.

AOS-0001 is therefore a starting point. Its success should be judged by whether it improves technical discussion, exposes missing responsibilities, enables interoperable implementations, and makes distributed agentic systems more governable and comprehensible. Planned specifications and aos-core are intended to test and refine the proposal through implementation, community review, and empirical evaluation.

# Appendix A. Canonical Data Objects

## A.1 IntentEnvelope

```
{
  "$schema": "urn:aos:schema:intent-envelope:1",
  "intent_id": "int-2026-0716-001",
  "version": 1,
  "created_at": "2026-07-16T17:12:00Z",
  "actor": {
    "type": "human",
    "id": "user-42",
    "identity_evidence": ["webauthn-credential-55"]
  },
  "objective": {
    "text": "Summarize the incident and recommend next actions",
    "language": "en",
    "objective_hash": "sha256:..."
  },
  "constraints": {
    "data_classification": "confidential",
    "allowed_regions": ["us-west"],
    "max_cost_usd": 2.00,
    "max_latency_ms": 5000,
    "deadline": "2026-07-16T18:00:00Z",
    "side_effects": "none"
  },
  "authority_ref": "auth-771",
  "policy_refs": ["enterprise-ai-policy-v12"],
  "context_refs": ["CASE-12345", "policy-snapshot-91"],
  "success_criteria": [
    {"type": "schema", "value": "urn:aos:schema:cited-summary:1"},
    {"type": "confidence", "operator": ">=", "value": 0.90}
  ],
  "trace_id": "4bf92f3577b34da6a3ce929d0e0e4736"
}
```

## A.2 CapabilityManifest

```
{
  "$schema": "urn:aos:schema:capability-manifest:1",
  "capability": {
    "id": "ai.case.summary.generate",
    "type": "workflow",
    "version": "1.0",
    "description": "Generate a structured incident summary with citations",
    "inputs": [
      {"name": "case_record", "schema": "urn:aos:schema:case-record:1"},
      {"name": "evidence", "schema": "urn:aos:schema:evidence-set:1"}
    ],
    "outputs": [
      {"name": "summary", "schema": "urn:aos:schema:cited-summary:1"}
    ],
    "side_effects": "none",
    "idempotency": "safe"
  },
  "provider": {
    "id": "internal-workflow-west",
    "owner": "example-enterprise",
    "implementation_version": "2026.07.3",
    "protocols": ["mcp", "http"],
    "regions": ["us-west"],
    "trust_profile": "trust-high",
    "confidence_profile": "summary-eval-v4",
```

```
    "latency_profile": "latency-west-v2",
    "cost_profile": "cost-medium",
    "compliance": ["soc2", "confidential-data"],
    "observability_profile": "full-lineage"
  }
}
```

## A.3 DelegationContext

```
{
  "$schema": "urn:aos:schema:delegation-context:1",
  "delegation_id": "del-002",
  "parent_delegation_id": "del-001",
  "root_intent_id": "int-2026-0716-001",
  "delegator": {"type": "agent", "id": "planner-agent-v3"},
  "delegatee": {"type": "agent", "id": "summary-agent-v1"},
  "capability_scope": ["ai.case.summary.generate"],
  "allowed_actions": ["read_case", "generate_summary"],
  "denied_actions": ["modify_case", "external_network"],
  "data_scope": ["CASE-12345"],
  "resource_budget": {"tokens": 12000, "cost_usd": 1.50},
  "obligations": ["full_lineage", "encrypted_temporary_state"],
  "valid_from": "2026-07-16T17:12:10Z",
  "valid_until": "2026-07-16T17:17:10Z",
  "revocation_ref": "revocation-channel-5",
  "trace_id": "4bf92f3577b34da6a3ce929d0e0e4736"
}
```

## A.4 ExecutionDirective

```
{
  "$schema": "urn:aos:schema:execution-directive:1",
  "execution_id": "exec-456",
  "intent_id": "int-2026-0716-001",
  "plan_id": "plan-77",
  "capability_id": "ai.case.summary.generate",
  "provider_id": "internal-workflow-west",
  "policy_decision_ref": "poldec-112",
  "trust_decision_ref": "trustdec-44",
  "delegation_context_ref": "del-002",
  "runtime_constraints": {
    "region": "us-west",
    "timeout_ms": 3500,
    "token_budget": 12000,
    "max_fanout": 3,
    "external_network": false,
    "sandbox_profile": "restricted-readonly"
  },
  "confidence_policy": {
    "required": 0.90,
    "on_low": ["fallback_provider", "human_escalation"]
  },
  "failure_policy": ["retry_once", "fallback_provider", "human_escalation"],
  "evidence_obligations": [
    "context-lineage", "tool-call-log", "provider-version", "output-hash"
  ]
}
```

## A.5 ConfidenceRecord

```
{
  "$schema": "urn:aos:schema:confidence-record:1",
  "confidence_id": "conf-88",
  "intent_id": "int-2026-0716-001",
  "execution_id": "exec-456",
  "components": {
    "intent": {"value": 0.96, "evidence": ["intent-validator-v2"]},
```

```
    "context": {"value": 0.88, "evidence": ["retrieval-eval-22"]},
    "data": {"value": 0.86, "evidence": ["freshness-check-11"]},
    "model": {"value": 0.84, "evidence": ["calibration-profile-4"]},
    "capability": {"value": 0.93, "evidence": ["provider-history-30d"]},
    "policy": {"value": 1.00, "evidence": ["policy-eval-112"]},
    "execution": {"value": 0.91, "evidence": ["schema-verifier-7"]},
    "temporal": {"value": 0.95, "evidence": ["clock-quality-3"]}
  },
  "aggregate": 0.91,
  "aggregation": {"method": "weighted_geometric_mean", "weights": "equal"},
  "uncertainty": {"lower": 0.87, "upper": 0.94},
  "decision": "proceed_with_warning",
  "reason": "model and data evidence below component warning threshold",
  "evaluator": "confidence-service-v2"
}
```

### A.6 Semantic Audit Event

```
{
  "$schema": "urn:aos:schema:audit-event:1",
  "event_id": "evt-901",
  "event": "capability.selected",
  "timestamp": "2026-07-16T17:12:11.123Z",
  "clock_uncertainty_ms": 0.8,
  "intent_id": "int-2026-0716-001",
  "execution_id": "exec-456",
  "delegation_id": "del-002",
  "parent_delegation_id": "del-001",
  "capability_id": "ai.case.summary.generate",
  "provider_id": "internal-workflow-west",
  "policy_decision": "allow_with_obligations",
  "trace_id": "4bf92f3577b34da6a3ce929d0e0e4736",
  "span_id": "00f067aa0ba902b7",
  "causal_parents": ["evt-897", "evt-899"],
  "evidence_refs": ["reg-20260716-1712", "policy-eval-112"],
  "integrity": {"hash": "sha256:...", "signature_ref": "sig-55"}
}
```

# Appendix B. Reference Sequences

## B.1 Normal Capability Invocation

1. A human or upstream system submits an IntentEnvelope.

2. The control plane validates identity and authority, normalizes the objective, and identifies applicable policy.

3. The capability registry returns candidate contracts and providers.

4. Hard constraints remove unauthorized, incompatible, unhealthy, or non-compliant candidates.

5. The control plane ranks feasible providers and creates an ExecutionDirective with delegation and evidence obligations.

6. The runtime invokes the provider through an adapter, applies context and runtime constraints, and records actual execution.

7. The result is verified, confidence is evaluated, and the control plane determines whether success criteria are met.

8. The caller receives the result and required lineage; audit evidence is persisted.

**End-to-End AOS Sequence Diagram**

*Human intent to system execution with governance, delegation, assurance, and audit*

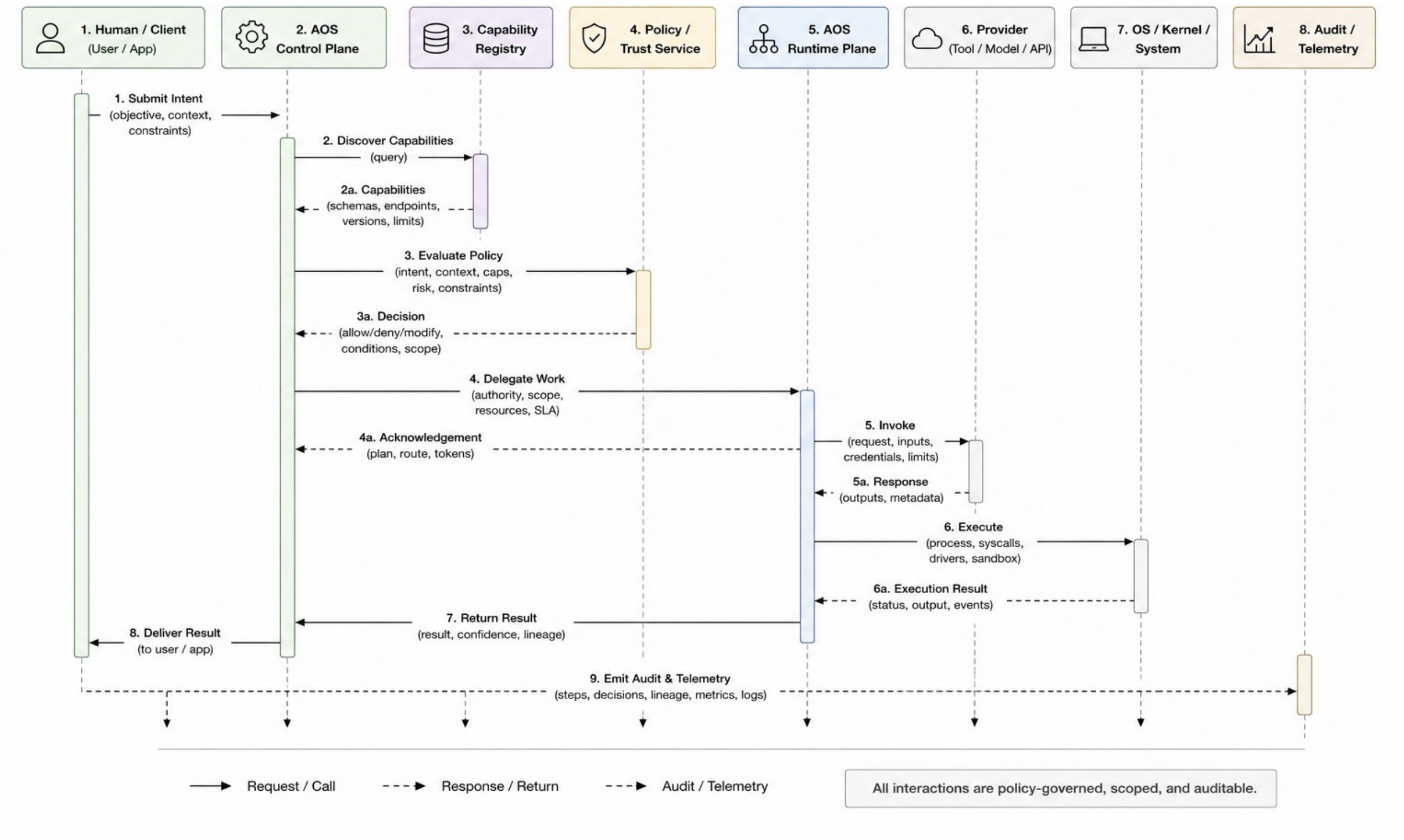

*Figure 13. End-to-end AOS sequence from human intent through discovery, policy and trust, runtime execution, provider and operating-system interaction, and audit telemetry.*

## B.2 Provider Failure and Regional Fallback

1. The primary provider exceeds a queue-delay or health threshold.

2. The runtime emits provider.degraded and requests fallback according to the directive.

3. The control plane checks that the next candidate remains feasible under current policy and data residency.

4. A new directive version selects the fallback provider or region and preserves the root intent and delegation lineage.

5. The runtime invokes the fallback with the remaining time, cost, and attempt budget.

6. The verifier compares the result to success criteria; exhausted options trigger human escalation or termination.

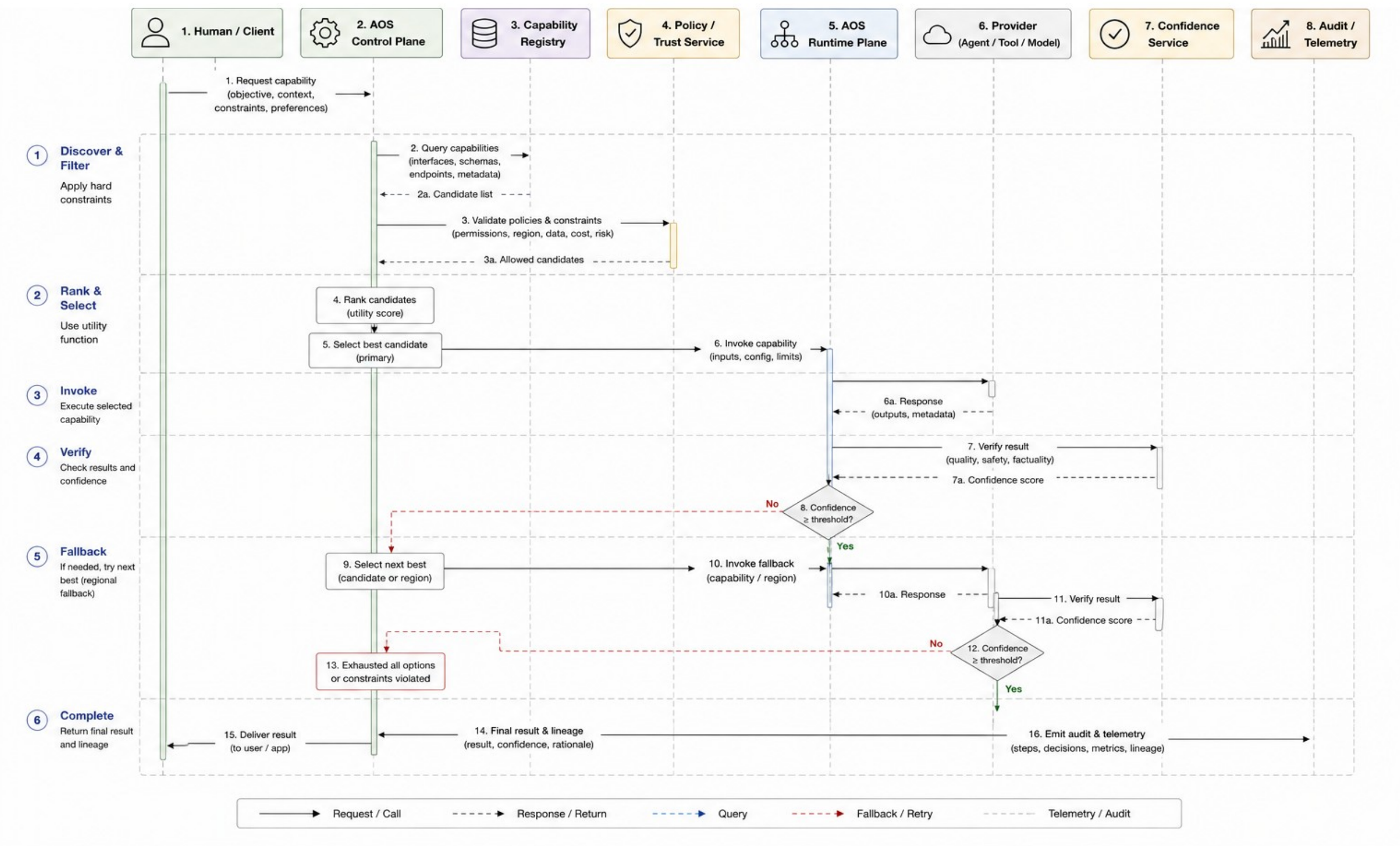

*Figure 14. Provider failure and regional fallback sequence with feasibility re-evaluation, bounded attempt budgets, confidence verification, and audit preservation.*

## B.3 Human Approval Sequence

1. Policy determines that the proposed action requires a human approver with a declared role.
2. The control plane assembles the objective, proposed plan, capability, predicted side effects, risk, confidence, and alternatives.
3. The runtime creates a human task with a deadline and an immutable review package.
4. The approver allows, denies, modifies, or requests additional evidence.
5. The decision is linked to the authority chain and becomes a new policy or approval record.
6. Execution proceeds only within the approved scope; timeout follows the safe default defined by policy.

## B.4 Revocation During Delegated Execution

1. The authority source revokes a parent delegation.
2. The control plane marks descendants invalid and emits revocation events to active runtimes.
3. The runtime stops new work, cancels or contains in-flight actions according to side-effect class, and reports unreachable downstream actors.
4. External credentials or leases are revoked where supported.
5. Audit records identify the time, propagation delay, actions completed before revocation, and residual exposure.
6. Human incident handling is triggered if the declared revocation bound is exceeded.

# Appendix C. Architecture Invariants and State Machines

## C.1 Foundational Invariants

*Table 16. Foundational AOS invariants.*

| Invariant | Requirement |
|---|---|
| **INV-001** | Execution MUST NOT occur without an originating intent or an explicitly declared system objective. |
| **INV-002** | A consequential action MUST NOT proceed without a valid authority chain and an applicable policy decision. |
| **INV-003** | Every provider invocation MUST realize a declared capability contract. |
| **INV-004** | Delegated scope, budget, and lifetime MUST NOT exceed the parent grant without independent authority. |
| **INV-005** | Hard policy constraints MUST be evaluated before utility optimization. |
| **INV-006** | Control decisions and runtime actions MUST be separately identifiable and jointly traceable. |
| **INV-007** | Externally visible lifecycle history MUST be monotonic; retries and replans MUST NOT erase failed attempts. |
| **INV-008** | Confidence decisions MUST retain evidence and MUST map to explicit control actions. |
| **INV-009** | Adapters and providers MUST NOT silently drop mandatory obligations. |
| **INV-010** | A conforming implementation MUST declare unsupported responsibilities and external dependencies. |

## C.2 Canonical Lifecycle

*Table 17. Canonical lifecycle states.*

| State | Meaning | Permitted next states |
|---|---|---|
| **RECEIVED** | Intent accepted and assigned an identifier | NORMALIZED, TERMINATED |
| **NORMALIZED** | Objective and constraints structured; ambiguity recorded | AUTHORIZED, ESCALATED, TERMINATED |
| **AUTHORIZED** | Authority and policy permit discovery and planning | DISCOVERING, ESCALATED, TERMINATED |
| **DISCOVERING** | Feasible capabilities and providers are evaluated | PLANNED, ESCALATED, TERMINATED, FAILED |
| **PLANNED** | Plan and execution directive prepared | EXECUTING, ESCALATED, TERMINATED |
| **EXECUTING** | Runtime is coordinating active work | VERIFYING, RETRYING, FALLBACK, ESCALATED, TERMINATED, FAILED |
| **VERIFYING** | Result and side effects are being assessed | COMPLETED, RETRYING, FALLBACK, ESCALATED, TERMINATED, FAILED |
| **RETRYING** | A bounded re-attempt is being prepared after a failed or unverified attempt | EXECUTING, FALLBACK, ESCALATED, TERMINATED, FAILED |
| **FALLBACK** | A feasible alternative provider or region is being selected and bound | EXECUTING, ESCALATED, TERMINATED, FAILED |
| **ESCALATED** | Work is paused pending an accountable human or independent-authority decision; non-terminal | AUTHORIZED, PLANNED, EXECUTING, TERMINATED, FAILED |
| **COMPLETED** | Success criteria satisfied and evidence finalized | Terminal |
| **TERMINATED** | Policy or authority deliberately stopped execution | Terminal |

| State | Meaning | Permitted next states |
|---|---|---|
| **FAILED** | Execution could not complete within permitted recovery | Terminal or new intent |

## C.3 Architectural Commitments of the AOS Series

This subsection summarizes the architectural commitments that anchor the AOS specification series. Each commitment is part of the current reference architecture and is maintained through the series' documented technical review process.

*Table 18. Architectural commitments of the AOS series.*

| ID | Proposed architectural commitment | Status | Basis |
|---|---|---|---|
| AC-001 | AOS contains both the Control & Governance Plane and the Runtime & Coordination Plane. | Proposed in AOS-0001 | Sections 8–10 |
| AC-002 | Platform services, conventional operating systems, and physical infrastructure remain outside the AOS boundary. | Proposed in AOS-0001 | Section 8 |
| AC-003 | Capability is the stable unit of discovery and provider substitution. | Proposed; builds on prior AOS work | Sections 6 and 9; Ref. [1] |
| AC-004 | Confidence is an actionable control signal rather than only an evaluation metric. | Proposed in AOS-0001 | Sections 6, 9, and 12 |
| AC-005 | Delegation and authority lineage are first-class architectural objects. | Proposed in AOS-0001 | Sections 6 and 9 |
| AC-006 | Observability retains semantic decision, policy, confidence, and delegation context. | Proposed in AOS-0001 | Sections 9 and 11 |
| AC-007 | AOS integrates existing runtimes and protocols rather than requiring their replacement. | Continued from prior AOS work | Sections 8, 10, and 11; Ref. [1] |
| AC-008 | Multi-region coordination may be federated; AOS does not require a globally centralized controller. | Open proposal | Sections 12 and 13 |
| AC-009 | Kernel-level assurance is an optional enforcement integration below the AOS semantic control boundary. | Open proposal | Sections 8, 14, and 16 |

# Appendix D. Threat Model

## D.1 Protected Assets

- Human and organizational authority;
- Intent integrity and success criteria;
- Capability registry and provider metadata;
- Policies, trust decisions, and confidence evidence;
- Context, memory, data, and credentials;
- Runtime state, budgets, and external side effects;
- Audit, provenance, and observability evidence.

## D.2 Representative Threats

The representative threats below draw on established security-control catalogs and adversarial threat taxonomies for AI systems [58, 59].

*Table 19. Representative AOS threat model.*

| Threat | Example | Architectural controls |
|---|---|---|
| **Intent manipulation** | Prompt injection, altered objective, hidden constraint removal | Immutable original intent, normalization provenance, policy validation, human clarification |
| **Registry poisoning** | False capability, malicious endpoint, stale metadata | Signed manifests, provenance, trust evaluation, health and revocation |
| **Authority escalation** | Agent or adapter expands delegated permissions | Monotonic delegation, short-lived grants, runtime enforcement, audit |
| **Policy bypass** | Direct provider invocation avoids control plane | Gateway enforcement, workload identity, network segmentation, conformance tests |
| **Context exfiltration** | Sensitive memory or retrieval sent to unauthorized provider | Data classification, provider feasibility, redaction, egress controls |
| **Tool abuse** | Authorized tool used with malicious arguments or excessive scope | Schema constraints, allowlists, side-effect metadata, sandboxing |
| **Delegation confusion** | Actions attributed to wrong delegation or actor | Delegation context propagation and delegation-scoped observability |
| **Confidence spoofing** | Provider inflates scores or suppresses uncertainty | Independent evaluators, calibration evidence, signed records, policy floors |
| **Retry amplification** | Failures cause unbounded calls, cost, or side effects | Attempt, fan-out, time, and cost budgets; idempotency; circuit breakers |
| **Audit tampering** | Logs dropped, reordered, or altered | Tamper-evident storage, causal links, sink acknowledgments, integrity proofs |
| **Kernel escape** | Runtime violates process or network constraints | Sandboxing, least privilege, seccomp/LSM/hypervisor controls, monitoring |
| **Human approval spoofing** | Fake approver or misleading review context | Strong identity, immutable review package, role policy, non-repudiation |

## D.3 Trust Boundaries

Trust boundaries may exist between human interfaces and the control plane, control and runtime planes, registries and providers, organizational domains, regions, platform services, host operating systems, and external networks. AOS profiles identify which identities and evidence cross each boundary.

A component inside the same process is not automatically trusted, consistent with zero-trust principles [49]. In-process plugins can manipulate memory or bypass interfaces. High-assurance profiles may isolate providers in separate processes or sandboxes and use signed, minimal contracts.

# Appendix E. Conformance Profiles

## E.1 Profile Model

A conformance profile defines a minimum set of AOS responsibilities, canonical objects, event semantics, security properties, and tests for a deployment class. Profiles prevent the architecture from becoming all-or-nothing. An embedded developer tool and a federated regulated system can both be AOS-conformant under different declared profiles.

A conformance claim names the AOS-0001 version, profile version, implemented responsibilities, external dependencies, protocol bindings, unsupported features, and test results.

*Table 20. Illustrative conformance profiles.*

| Profile | Minimum scope | Key distinctions |
|---|---|---|
| **Embedded Control** | One host or process; local registry; basic intent, policy, capability, confidence, audit | Federation, cross-region scheduling, kernel assurance optional |
| **Enterprise Control** | Central or regional control; multiple runtimes; identity, policy, delegation, full semantic telemetry | Declared model, tool, workflow, and platform adapters |
| **Federated Enterprise** | Multiple administrative domains; policy and registry federation; signed evidence and revocation | Cross-domain trust, consistency, and data-sharing profile |
| **Sovereign** | Isolated authority, models, data, registry, and evidence | No undeclared external provider or telemetry dependency |
| **High Assurance** | Independent verification, deterministic enforcement, complete lineage, formal invariants | Stronger identity, runtime isolation, evidence integrity, and human approval |

## E.2 Example Conformance Statement

```
Implementation: aos-core 0.x
AOS-0001 version: draft-0.8
Profile: Embedded Control (experimental)
Implemented:
  - intent normalization and validation
  - capability registry, matching, and ranking
  - policy and trust-aware routing
  - delegation context and lineage
  - confidence decisions
  - audit and observability events
  - MCP, A2A, HTTP, and local execution adapters
External dependencies:
  - Python runtime
  - selected policy, identity, model, tool, and telemetry providers
Unsupported or experimental:
  - federated registry
  - cross-region scheduler
  - kernel assurance
  - production durability and high availability
```

# Appendix F. Glossary

*Table 21. AOS glossary.*

| Term | Definition |
|---|---|
| **Actor** | A human, agent, service, device, organization, or runtime identity participating in AOS. |
| **Agent** | A goal-directed computational actor that can reason, plan, invoke capabilities, and maintain state. |
| **AOS domain** | An administrative and trust boundary operating one or more AOS components. |
| **Assurance** | Evidence-based evaluation that behavior remains within authorized and expected bounds. |
| **Authority** | The right to request, approve, delegate, or execute an action under specified scope. |
| **Capability** | A stable contract describing an outcome that a provider can realize. |
| **Confidence** | A system-level assurance signal linked to evidence and a control action. |
| **Consequential action** | An action with declared non-none side effects, external state modification, resource commitment, or other irreversible impact; the trigger for INV-002 and human-oversight requirements. The full side-effect taxonomy is deferred to AOS-0004. |
| **Context** | Bounded operational state selected for a decision or execution step. |
| **Control plane** | AOS plane that interprets intent and creates authoritative governance decisions. |
| **Delegation** | Bounded transfer of authority with lineage, constraints, and revocation. |
| **Evidence** | Telemetry, decisions, attestations, validations, artifacts, and lineage used to justify or reconstruct behavior. |
| **Execution directive** | Authorized contract from the control plane to the runtime plane. |
| **Intent** | Desired outcome plus constraints, authority, context references, success criteria, and temporal requirements. |
| **Obligation** | A condition that must be satisfied during or after an allowed action. |
| **Plan** | Reviewable decomposition of intent into steps, dependencies, checkpoints, and failure actions. |
| **Policy** | Machine-evaluable rule controlling eligibility, constraints, approvals, obligations, and response. |
| **Provider** | Concrete implementation of one or more capabilities. |
| **Runtime plane** | AOS plane that coordinates approved execution and emits evidence. |
| **Semantic observability** | Telemetry enriched with intent, policy, trust, confidence, capability, context, delegation, and causality. |
| **Trust** | Contextual, evidence-backed acceptability of an actor, provider, artifact, or environment. |

# Appendix G. Relationship to Prior AOS Work and the Planned Specification Series

The AOS concept was initially introduced by Sharma and Shah in "Agent Operating Systems (AOS): Integrating Agentic Control Planes into, and Beyond, Traditional Operating Systems" [1]. That work defined the initial AOS motivation, assumptions, non-goals, system responsibilities, integration models, Linux and Windows mappings, security implications, and evaluation criteria.

The present manuscript is a direct continuation of that work rather than an independent or competing definition. AOS-0001 extends the earlier control-plane-oriented concept into a vendor-neutral two-plane reference operating architecture. It formalizes the AOS boundary, separates the Control & Governance Plane from the Runtime & Coordination Plane, distinguishes external platform and operating-system dependencies, and introduces foundational concepts, interfaces, invariants, mathematical models, deployment profiles, and conformance guidance.

Only the earlier AOS paper cited as [1] has been published at the time of this draft. AOS-0001 is the current manuscript. AOS-0002 through AOS-0010 are planned documents and must not be cited as published specifications. Their titles and scopes remain subject to revision through the AOS editorial and technical-review process.

*Table 22. Relationship of published, current, and planned AOS documents.*

| Identifier | Status | Relationship to AOS-0001 |
| --- | --- | --- |
| AOS-0000 | Published as arXiv:2606.01508 [1] | Introduces the original AOS concept, assumptions, responsibilities, operating-system integration models, and security considerations. |
| AOS-0001 | Current draft | Defines the foundation-level, two-plane AOS reference operating architecture. |
| AOS-0002 | Planned; not yet published | Will formalize plane and layer boundaries, profiles, terminology, and conformance vocabulary. |
| AOS-0003 | Planned; not yet published | Will specify the Control & Governance Plane in normative detail. |
| AOS-0004 | Planned; not yet published | Will define capability contracts, manifests, registration, discovery, negotiation, ranking, and federation. |
| AOS-0005 | Planned; not yet published | Will define confidence evidence, aggregation, propagation, control actions, and runtime assurance. |
| AOS-0006 | Planned; not yet published | Will define end-to-end reliability dimensions, controls, metrics, and evaluation methods. |
| AOS-0007 | Planned; not yet published | Will map protocols and ecosystem technologies to AOS interfaces and responsibilities. |
| AOS-0008 | Planned; not yet published | Will define the open AOS landscape and classification taxonomy. |
| AOS-0009 | Planned; not yet published | Will define the enterprise AOS maturity and assessment model. |
| AOS-0010 | Planned; implementation evolving | Will document aos-core as one non-exclusive reference implementation and carry the executable conformance examples. |

Planned specifications may refine or specialize AOS-0001, but they must not silently redefine its foundational boundaries. A proposal that contradicts AOS-0001 must identify the affected architectural commitment, provide a technical rationale, describe compatibility and migration implications, and result in an explicitly versioned revision of AOS-0001.

The AOS landscape, maturity model, conformance suite, and aos-core implementation are supporting artifacts. They may provide evidence and implementation feedback, but they do not independently redefine the reference architecture.

# Acknowledgments

This work was conducted as part of the Unified Intelligent Infrastructure workstream at the Open Compute Project (OCP), https://www.opencompute.org/w/index.php?title=Unified_Intelligent_Infrastructure.

# References

[1] A. Sharma and D. Shah, “Agent Operating Systems (AOS): Integrating Agentic Control Planes into, and Beyond, Traditional Operating Systems,” arXiv:2606.01508, 2026.
[2] K. Mei et al., “AIOS: LLM Agent Operating System,” arXiv:2403.16971, 2024.
[3] H. Huang et al., “TopoClaw: A Human-Centric and Topology-Aware Agent Operating System,” arXiv:2605.15556, 2026.
[4] A. Mishra and K. Sharad, “Observability for Delegated Execution in Agentic AI Systems,” arXiv:2606.09692, 2026.
[5] E. Tabassi, “Artificial Intelligence Risk Management Framework (AI RMF 1.0),” NIST AI 100-1, National Institute of Standards and Technology, 2023.
[6] C. Autio et al., “Artificial Intelligence Risk Management Framework: Generative Artificial Intelligence Profile,” NIST AI 600-1, National Institute of Standards and Technology, 2024.
[7] Model Context Protocol, “Model Context Protocol Specification, revision 2025-11-25,” 2025.
[8] A2A Project, “Agent2Agent Protocol Specification and Documentation,” Linux Foundation project, accessed 2026.
[9] OpenTelemetry, “OpenTelemetry Concepts and Signals,” documentation, accessed 2026.
[10] W3C, “Trace Context,” W3C Recommendation, 23 November 2021.
[11] Kubernetes Authors, “Kubernetes Components and Cluster Architecture,” Kubernetes Documentation, accessed 2026.
[12] S. Bradner, “Key words for use in RFCs to Indicate Requirement Levels,” RFC 2119, 1997.
[13] B. Leiba, “Ambiguity of Uppercase vs Lowercase in RFC 2119 Key Words,” RFC 8174, 2017.
[14] L. Lamport, “Time, Clocks, and the Ordering of Events in a Distributed System,” Communications of the ACM, vol. 21, no. 7, pp. 558–565, 1978.
[15] L. Lamport, “The Part-Time Parliament,” ACM Transactions on Computer Systems, vol. 16, no. 2, pp. 133–169, 1998.
[16] B. H. Sigelman et al., “Dapper, a Large-Scale Distributed Systems Tracing Infrastructure,” Google Research, 2010.
[17] J. Dean and L. A. Barroso, “The Tail at Scale,” Communications of the ACM, vol. 56, no. 2, pp. 74–80, 2013.
[18] A. Burns and A. Wellings, Real-Time Systems and Programming Languages, 4th ed., Addison-Wesley, 2009.
[19] J. L. Hennessy and D. A. Patterson, Computer Architecture: A Quantitative Approach, 6th ed., Morgan Kaufmann, 2017.
[20] A. Silberschatz, P. B. Galvin, and G. Gagne, Operating System Concepts, 10th ed., Wiley, 2018.
[21] A. S. Tanenbaum and H. Bos, Modern Operating Systems, 4th ed., Pearson, 2014.
[22] Open Policy Agent, “OPA Documentation and Policy Language,” Cloud Native Computing Foundation project, accessed 2026.
[23] SPIFFE, “SPIFFE Standards and Workload Identity Specification,” Cloud Native Computing Foundation project, accessed 2026.
[24] Linux Kernel Documentation, “Seccomp BPF (Secure Computing with Filters),” accessed 2026.
[25] T. Heo, “Control Group v2,” Linux Kernel Documentation, accessed 2026.
[26] W3C, “PROV-O: The PROV Ontology,” W3C Recommendation, 2013.
[27] L. Torres-Arias et al., “in-toto: Providing Farm-to-Table Guarantees for Bits and Bytes,” USENIX Security Symposium, 2019.
[28] OpenSSF, “Supply-chain Levels for Software Artifacts (SLSA),” specification, accessed 2026.
[29] Y. Bai et al., “Constitutional AI: Harmlessness from AI Feedback,” arXiv:2212.08073, 2022.
[30] G. Irving, P. Christiano, and D. Amodei, “AI Safety via Debate,” arXiv:1805.00899, 2018.
[31] L. Huang et al., “A Survey on Hallucination in Large Language Models: Principles, Taxonomy, Challenges, and Open Questions,” arXiv:2311.05232, 2023.
[32] S. Yao et al., “ReAct: Synergizing Reasoning and Acting in Language Models,” arXiv:2210.03629, 2022.
[33] T. Schick et al., “Toolformer: Language Models Can Teach Themselves to Use Tools,” Advances in Neural Information Processing Systems, 2023.
[34] Q. Wu et al., “AutoGen: Enabling Next-Gen LLM Applications via Multi-Agent Conversation,” arXiv:2308.08155, 2023.
[35] LangChain, “LangGraph Documentation,” accessed 2026.
[36] CrewAI, “CrewAI Documentation,” accessed 2026.
[37] P. Moritz et al., “Ray: A Distributed Framework for Emerging AI Applications,” USENIX OSDI, 2018.
[38] D. G. Murray et al., “Naiad: A Timely Dataflow System,” ACM SOSP, 2013.
[39] A. Verma et al., “Large-Scale Cluster Management at Google with Borg,” EuroSys, 2015.
[40] M. Schwarzkopf et al., “Omega: Flexible, Scalable Schedulers for Large Compute Clusters,” EuroSys, 2013.
[41] B. Hindman et al., “Mesos: A Platform for Fine-Grained Resource Sharing in the Data Center,” USENIX NSDI, 2011.
[42] K. J. Åström and R. M. Murray, Feedback Systems: An Introduction for Scientists and Engineers, Princeton University Press, 2008.
[43] L. Kleinrock, Queueing Systems, Volume I: Theory, Wiley, 1975.
[44] S. Gilbert and N. Lynch, “Brewer’s Conjecture and the Feasibility of Consistent, Available, Partition-Tolerant Web Services,” SIGACT News, 2002.
[45] M. J. Fischer, N. A. Lynch, and M. S. Paterson, “Impossibility of Distributed Consensus with One Faulty Process,” Journal of the ACM, 1985.
[46] R. Schwartz et al., “Green AI,” Communications of the ACM, vol. 63, no. 12, pp. 54–63, 2020.
[47] S. Amershi et al., “Guidelines for Human-AI Interaction,” ACM CHI, 2019.
[48] V. Vovk, A. Gammerman, and G. Shafer, Algorithmic Learning in a Random World, Springer, 2005.
[49] NIST, “Zero Trust Architecture,” NIST SP 800-207, 2020.
[50] Cloud Native Computing Foundation, “CloudEvents Specification,” accessed 2026.
[51] D. Hardt, “The OAuth 2.0 Authorization Framework,” RFC 6749, 2012.
[52] OpenID Foundation, “OpenID Connect Core 1.0,” 2014.

[53] W3C, “Web Authentication: An API for accessing Public Key Credentials (WebAuthn),” W3C Recommendation, accessed 2026.
[54] ISO/IEC 42001:2023, Information technology — Artificial intelligence — Management system.
[55] ISO/IEC 23894:2023, Information technology — Artificial intelligence — Guidance on risk management.
[56] B. Beyer et al., Site Reliability Engineering: How Google Runs Production Systems, O’Reilly Media, 2016.
[57] L. Lamport, Specifying Systems: The TLA+ Language and Tools for Hardware and Software Engineers, Addison-Wesley, 2002.
[58] NIST, “Security and Privacy Controls for Information Systems and Organizations,” NIST SP 800-53 Rev. 5, 2020.
[59] MITRE, “Adversarial Threat Landscape for Artificial-Intelligence Systems (ATLAS),” knowledge base, accessed 2026.
[60] A. D. Kshemkalyani and M. Singhal, Distributed Computing: Principles, Algorithms, and Systems, Cambridge University Press, 2008.
[61] J. B. Dennis and E. C. Van Horn, "Programming Semantics for Multiprogrammed Computations," Communications of the ACM, vol. 9, no. 3, pp. 143–155, 1966.
[62] M. S. Miller, "Robust Composition: Towards a Unified Approach to Access Control and Concurrency Control," Ph.D. dissertation, Johns Hopkins University, 2006.
[63] N. Hardy, "The Confused Deputy (or Why Capabilities Might Have Been Invented)," ACM SIGOPS Operating Systems Review, vol. 22, no. 4, pp. 36–38, 1988.
[64] A. Birgisson et al., "Macaroons: Cookies with Contextual Caveats for Decentralized Authorization in the Cloud," Network and Distributed System Security Symposium (NDSS), 2014.
[65] Foundation for Intelligent Physical Agents, "FIPA Agent Management Specification," SC00023K, 2004.
[66] OASIS, "UDDI Version 3.0.2," OASIS Standard, 2004.